\documentclass{article} %
\usepackage{iclr2027_conference,times}

\usepackage{hyperref}
\usepackage{url}

\usepackage[utf8]{inputenc} %
\usepackage[T1]{fontenc}
\usepackage{booktabs}       %
\usepackage{amsfonts}       %
\usepackage{nicefrac}       %
\usepackage{microtype}      %
\usepackage{xcolor}         %
\usepackage{wrapfig}
\usepackage{natbib}
\usepackage{subcaption}
\usepackage{multirow}
\usepackage{longtable}
\usepackage{algorithm}
\usepackage{algpseudocode}
\usepackage{makecell}

\def\eqref#1{(\ref{#1})}

\usepackage{amsmath}
\usepackage{amssymb}
\usepackage{mathtools}
\usepackage{amsthm}

\usepackage{amsmath,amsfonts,bm}

\def\eqref#1{equation~\ref{#1}}

\def\1{\bm{1}}

\def\vh{{\bm{h}}}

\def\vx{{\bm{x}}}

\DeclareMathAlphabet{\mathsfit}{\encodingdefault}{\sfdefault}{m}{sl}
\SetMathAlphabet{\mathsfit}{bold}{\encodingdefault}{\sfdefault}{bx}{n}

\newcommand{\punt}[1]{}

\usepackage{amsthm}
\usepackage{xcolor}
\usepackage{hyperref}

\renewcommand{\eqref}[1]{eq.~\ref{eq:#1}}
\newcommand{\Nrm}{\mathcal{N}}

\newcommand{\tabref}[1]{Table~\ref{tab:#1}}  %

\newtheorem{thm}{Theorem}[section]

\newtheorem{prop}[thm]{Proposition}

\def\Pr{\ensuremath{\text{Pr}}}

\title{Privacy Accounting for Empirical Denoisers}

\author{William Cheng \\
University of British Columbia \\
  \texttt{willc135@cs.ubc.ca} \\
  \And
  Dongjun Kim \thanks{Now at Adobe.}\\
  Stanford University \\
  \texttt{dongjun@stanford.edu} \\
  \AND
  Mijung Park \\
  University of British Columbia \\
  \texttt{mijungp@cs.ubc.ca} \\
}

\iclrfinalcopy %
\begin{document}

\maketitle

\begin{abstract}
Diffusion models have emerged as the foundation of modern generative systems, yet their high memorization capacity raises privacy concerns. While differentially private (DP) training provides formal guarantees, it comes at a cost in sample quality.
In this work, we study a complementary question: \textit{how much privacy loss is incurred during the sampling process itself?} Directly analyzing modern neural denoisers is challenging because their sensitivity to a single training example is generally intractable. To overcome this challenge, we construct and analyze a  \textit{clipped privacy-accountable empirical denoiser}, a training-free estimator of the Bayes-optimal denoiser, whose per-step sensitivity can be computed explicitly, allowing each denoising step to be a Gaussian mechanism. This perspective enables privacy accounting under Gaussian DP when each training example contributes to each denoising step, and under $f$-DP trade-off functions when each generated sample uses a single randomly subsampled block of the training data. We report every $(\varepsilon, \delta)$ guarantee as a certified upper bound from a privacy loss distribution accountant, and use a cumulant-based Edgeworth approximation as a fast diagnostic, characterizing where it is reliable and where it becomes anti-conservative. We also analyze the privacy loss along the denoising trajectory and show that it grows steeply near the critical window in which semantic content is formed. This observation motivates hybrid samplers that use the private empirical denoiser through the formative steps where it remains a faithful proxy for the neural denoiser, then switch to a public neural denoiser for the remaining refinement. Our results on CIFAR-10 and CelebA-64 show that the privacy of diffusion models depends not only on the training procedure, but also on sampling-time design choices such as subsampling, stopping time, and hybrid denoising.
\end{abstract}

\section{Introduction}
Diffusion models~\citep{sohldickstein2015deep,DDPM} are now central to modern generative systems, but their ability to reproduce training examples raises serious privacy concerns~\citep{wu2023membership,carlini2023extracting,tang2023membership,hu2023membership,duan2023are,Matsumoto23MembershipInference}. Post-hoc filters and memorization-aware model changes can reduce visible leakage, but they do not give attack-agnostic privacy guarantees. Differentially private (DP) training gives such guarantees, but it answers a different question: it bounds what the weights reveal, while users and regulators see only generated samples. Sampling-time leakage is a separate surface that persists even when the denoiser is trained non-privately, whether or not DP training is practical at scale. This raises a complementary question: \textit{how much privacy loss is incurred by the sampling process itself?}

Answering this requires treating the sampler, rather than the training algorithm, as the mechanism to be analyzed. The obstacle is that a neural denoiser has intractable data sensitivity, so its denoising steps admit no explicit privacy accounting. To make the problem analyzable, we privatize an \emph{empirical denoiser}~\citep{niedoba2024towards} whose sensitivity can be computed directly. Since each reverse diffusion step already injects Gaussian noise, the empirical denoising update can be interpreted as a Gaussian mechanism, allowing us to track the sampling trajectory with Gaussian Differential Privacy (GDP) \citep{GDP}.

However, the GDP composition using the full batch of training data yields a poor privacy level. Denoising with subsampled data at each denoising step can improve the privacy level. However, this hurts the diffusion sampling, as 
every newly drawn subset of data injects extra stochastic drift into the sampling trajectory, which prevents coherent semantics from forming across the trajectory. To this end, we propose \textit{output-wise subsampling}, subsample a block from the dataset once and reuse it for the entire denoising trajectory to produce one sample. This stabilizes the denoising direction and achieves a decent privacy level. 

When converting $f$-DP to $(\epsilon, \delta)$-DP, we encounter unreliable conversion, especially at small $\delta$, when using the CLT-type $f$-DP accounting that assumes the CDF of the standardized test statistic under the composition is simply Gaussian. We report all guarantees as certified upper bounds from a PLD accountant~\citep{doroshenko2022connect}, and study a cumulant-based \textit{Edgeworth approximation}~\citep{zheng2020sharp} as a fast estimate and measure where it is accurate and where it understates the certified value.

Our contributions are summarized below:
\begin{enumerate}
    \item We privatize a clipped empirical denoiser with explicit per-step sensitivity of $2C/|\mathcal{D}|$. Diffusion sampling then becomes a composition of Gaussian mechanisms, with no model trained on the private data.
    \item We derive privacy guarantees for full-batch and output-wise subsampled diffusion sampling using GDP and $f$-DP trade-off functions, and report certified $(\epsilon,\delta)$ upper bounds from a PLD accountant. We use the Edgeworth approximation as a diagnostic where cumulant checks pass, and characterize where it fails. Subsampling amplification applies once per reused block, rather than once per step.
    \item We show that the per-step privacy cost grows along the trajectory relative to the critical window, measure where the empirical denoiser is a window-local semantic proxy for a neural denoiser, and use both to place the private-to-public handoff,
    evaluated on CIFAR-10 and CelebA-64.
\end{enumerate}

\section{Background}
\subsection{Denoising Diffusion Models}

Diffusion models are built upon the forward and reverse processes \citep{DDPM, song2020score}. The forward perturbation describes how the noise is gradually corrupting data sample $\mathbf{x}_{0}$, defined as $\mathbf{x}_{t}=\alpha_{t}\mathbf{x}_{0}+\sigma_{t}\bm{\epsilon}$, where $\bm{\epsilon}$ is a standard Gaussian variable, $\alpha_{t}$ and $\sigma_{t}$ are prescribed time-dependent mean and variance coefficients, respectively. For simplicity, we describe the diffusion process from EDM~\citep{karras2022elucidating}, but our framework is adaptable to any diffusion processes such as DDPM~\citep{DDPM} or Flow Matching (FM)~\citep{lipman2022flow}. In EDM, the forward diffusion process is simply defined by $\mathbf{x}_{t}=\mathbf{x}_{0}+t\bm{\epsilon}$, i.e., $\alpha_{t}\equiv 1$ and $\sigma_{t}=t$, and the reverse denoising process is computed by
\begin{align}\label{eq:denoising}
    \mathbf{x}_{t-\Delta t}=\left(1-2\frac{\Delta t}{t}\right)\mathbf{x}_{t}+2\frac{\Delta t}{t}\mathbb{E}[\mathbf{x}\vert\mathbf{x}_{t}]+\sqrt{2t\Delta t}\bm{\epsilon}.
\end{align}
The term $\mathbb{E}[\mathbf{x}\vert\mathbf{x}_{t}]$, called a Bayes optimal denoiser, is the conditional expectation of the clean data given the noisy observation $\mathbf{x}_{t}$. In practice, a neural denoiser $D(\mathbf{x}_{t},t;\bm{\theta}(\mathcal{D}))$ approximates this quantity, where $\bm{\theta}(\mathcal{D})$ are parameters learned from dataset $\mathcal{D}$. By rolling out \eqref{denoising} up to time zero, diffusion model gradually synthesizes a sample.

\subsection{Differential Privacy and Gaussian Mechanism}

Differential Privacy (DP) provides a formal framework to limit the influence of any single record on the output of a randomized algorithm. Intuitively, whether or not an individual’s data is present, the distribution of outputs should remain (almost) unchanged. This view naturally extends to diffusion sampling, which is itself a randomized mechanism.

A fundamental building block is the \emph{Gaussian mechanism}: given a function $h:\mathcal{D}\mapsto\mathbb{R}^p$, the mechanism releases
\begin{align}\label{eq:gaussian_mechanism}
   \mathcal{M}_{\mathrm{Gauss}}(\mathcal{D};h)
   = h(\mathcal{D}) + \bm{n}, 
   \qquad \bm{n}\sim \mathcal{N}(0,\sigma^{2}\Delta_{h}^{2}\mathbf{I}),
\end{align}
where $\Delta_{h}$ is the global sensitivity—the maximum change in $h$ between neighboring datasets. By calibrating the noise to $\Delta_{h}$, the Gaussian mechanism guarantees DP (see Appendix~\ref{sec:appendix_dp} for formal definitions).

For analyzing multi-step procedures such as diffusion sampling, we adopt \emph{Gaussian Differential Privacy (GDP)}~\citep{GDP}. GDP tightly characterizes the privacy loss of Gaussian mechanisms and, crucially, admits an exact composition rule:
\begin{align}\label{eq:gdp_composition}
   \mu = \sqrt{\sum_{i} \mu_i^2},
\end{align}
where $\mu_i$ is the GDP parameter of the $i$-th step. For example, in the case of \eqref{gaussian_mechanism} with $\bm{n}\sim\mathcal{N}(0,\sigma^{2}\Delta_{\vh}^{2}\mathbf{I})$, the Gaussian mechanism corresponds to $\mu_i$-GDP with $\mu_i = \frac{1}{\sigma}$. This makes GDP especially well-suited for diffusion processes, which inject Gaussian noise at every denoising step.

Two further facts are important for our analysis. First, privacy amplification by subsampling can further reduce the effective $\mu$. Second, GDP parameters can be translated into $(\epsilon,\delta)$ guarantees when needed for regulatory reporting; we provide details in Appendix~\ref{sec:appendix_gdp}. Together, these tools allow us to connect the denoising process directly to established DP accounting methods.

\subsection{Privacy Loss Distributions}
\label{subsec:pld}

$(\epsilon,\delta)$-DP is equivalently characterized by the hockey-stick divergence: a mechanism is $(\epsilon,\delta)$-DP iff $H_{e^\epsilon}(M(\mathcal{D})\,\|\,M(\mathcal{D}'))\le\delta$ for all neighboring $\mathcal{D},\mathcal{D}'$, where $H_{\gamma}(P\|Q)=\int [P(x)-\gamma Q(x)]_+\,dx$. The \emph{privacy loss distribution} (PLD) is the law of the log-likelihood ratio $\log\frac{P(x)}{Q(x)}$ under $P$; the hockey-stick divergence, and hence $\delta(\epsilon)$, is a functional of it, and the PLD of a composition is the convolution of the individual PLDs. Numerical PLD accountants evaluate this convolution on a discretized support~\citep{gopi2021numerical,doroshenko2022connect}. Discretization can be performed \emph{pessimistically}, overestimating $H_{e^\epsilon}$ for every $\epsilon$ and hence yielding a valid $(\epsilon,\delta)$ upper bound, or \emph{optimistically}, underestimating it. We use the pessimistic estimate for all reported guarantees, and the optimistic estimate to quantify how much of the reported $\epsilon$ is discretization slack rather than true privacy loss. This distinction is what we mean by \emph{certified} throughout: the reported value is an upper bound by construction, not an approximation of unknown sign.

\section{Privacy Analysis of Denoising Process with Empirical Denoiser}\label{sec:method}
\subsection{Single-Step Denoising is a GDP}

The Bayes denoiser in \eqref{denoising} is the conditional expectation $\mathbb{E}[\mathbf{x}_0\vert\mathbf{x}_t]=p_t(\mathbf{x}_t)^{-1}\int \mathbf{x}_0\,p(\mathbf{x}_t\vert\mathbf{x}_0)\,p(\mathbf{x}_0)\,d\mathbf{x}_0$, where $p(\mathbf{x}_t\mid\mathbf{x}_0)=\mathcal{N}(\mathbf{x}_t;\mathbf{x}_0,t^2 I)$ is the closed-form EDM forward transition density and $p_t(\mathbf{x}_t)=\int p(\mathbf{x}_t\mid\mathbf{x}_0)\,p(\mathbf{x}_0)\,d\mathbf{x}_0$ is the marginal of the noisy state. Real-world samplers approximate this Bayes denoiser with a neural denoiser $D(\mathbf{x}_t,t;\bm{\theta}(\mathcal{D}))$, but the sensitivity of $D$ to changing one training example is not tractable, so it admits no direct privacy accounting. We therefore analyze the empirical denoiser, which replaces $p(\mathbf{x}_0)$ with the empirical measure over $\mathcal{D}$ and uses the empirical marginal $\widehat{p}_t(\mathbf{x}_t)=\tfrac{1}{|\mathcal{D}|}\sum_{\mathbf{x}_0\in\mathcal{D}}p(\mathbf{x}_t\mid\mathbf{x}_0)$ as its normalizer. Unlike the transition kernel, $\widehat{p}_t$ is data-dependent and cannot enter a private mechanism without its own sensitivity accounting.

\begin{equation}\label{eq:empirical}
\hat{\mathbb{E}}[\mathbf{x}\vert\mathbf{x}_{t};\mathcal{D}]
=
\frac{1}{\widehat{p}_t(\mathbf{x}_t)|\mathcal{D}|}
\sum_{\mathbf{x}_0\sim\mathcal{D}}\mathbf{x}_0p(\mathbf{x}_t\vert\mathbf{x}_0).
\end{equation}

Substituting \eqref{empirical} into \eqref{denoising} gives
\begin{equation}\label{eq:combined_equation}
\mathbf{x}_{t-\Delta t} = \Big(1-\tfrac{2\Delta t}{t}\Big)\mathbf{x}_t
+ \tfrac{2\Delta t}{t}\underbrace{\tfrac{1}{p_t(\mathbf{x}_t)|\mathcal{D}|}\textstyle\sum_{\mathbf{x}_0\in\mathcal{D}}\mathbf{x}_0 p(\mathbf{x}_t|\mathbf{x}_0)}_{\text{data-dependent}} + \sqrt{2t\Delta t}\,\epsilon ,
\end{equation}
so the only term depending on $\mathcal{D}$ is the empirical average, and the noise $\sqrt{2t\Delta t}\,\epsilon$ already injected by the sampler can serve as the noise of a Gaussian mechanism once this average has bounded sensitivity. Let $\mathbf{v}(\mathbf{x}_{0};\mathbf{x}_{t})=\mathbf{x}_{0}p(\mathbf{x}_{t}\vert\mathbf{x}_{0})$ and let $\bar{\mathbf{v}}(\mathbf{x}_{0};\mathbf{x}_{t})=\mathbf{v}(\mathbf{x}_{0};\mathbf{x}_{t})/\max(1,\Vert\mathbf{v}(\mathbf{x}_{0};\mathbf{x}_{t})\Vert/C)$ be its clipped version, so that $\Vert\bar{\mathbf{v}}\Vert\le C$.
Computing $\widehat{p}_t(\mathbf{x}_t)$ privately would require an additional data-dependent pass, so we replace it with a public constant $\beta$ ($\beta=1$ in experiments). This yields a deliberately \emph{unnormalized} empirical mechanism. $\beta=1$ is a design choice, not a claim that $\widehat{p}_t(\mathbf{x}_t)\approx 1$, and $\beta$ is a genuine privacy--utility parameter that we ablate in Appendix~\ref{app:beta-ablation}. Substituting the clipped average into \eqref{denoising} gives
\begin{align*}
\mathbf{x}_{t-\Delta t} = \Big(1-\tfrac{2\Delta t}{t}\Big)\mathbf{x}_t + \tfrac{2\Delta t}{t}\,\tfrac{1}{\beta|\mathcal{D}|}\textstyle\sum_{\mathbf{x}_0\in\mathcal{D}}\bar{\mathbf{v}}(\mathbf{x}_0;\mathbf{x}_t) + \sqrt{2t\Delta t}\,\bm{\epsilon},
\end{align*}
and the public invertible rescaling by $\beta t/(2\Delta t)$ shows that the data-dependent part of one step is the Gaussian mechanism
\begin{align}\label{eq:norm_clipped}
   \mathcal{M}_{\mathrm{Gauss}}(\mathcal{D};\mathbf{x}_t)
   =
   \frac{1}{|\mathcal{D}|}\sum_{\mathbf{x}_{0}\in \mathcal{D}}
      \bar{\mathbf{v}}(\mathbf{x}_{0};\mathbf{x}_{t})
   +
   \beta
   \sqrt{\frac{t^{3}}{2\Delta t}}\;\bm{\epsilon}.
\end{align}

For neighboring datasets, where $\mathcal{D}'$ adds or removes one record of $\mathcal{D}$, the clipped average has sensitivity at most $2C/|\mathcal{D}|$, so \eqref{norm_clipped} is $\mu_t$-GDP with $\mu_t=\frac{2C}{\beta|\mathcal{D}|}\sqrt{2\Delta t/t^3}$. This bound is conservative in the sense that it treats the empirical denoising direction as the private release at every step and accounts for the sampling noise already injected by the diffusion process.

\begin{figure}[t]
    \centering
    \begin{minipage}[c]{0.4\linewidth}
        \centering
        \includegraphics[width=\linewidth]{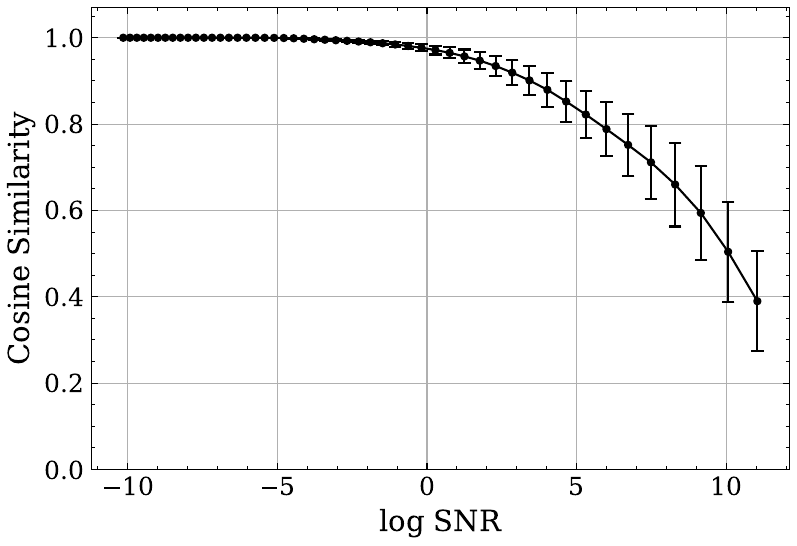}
    \end{minipage}
    \hfill
    \begin{minipage}[c]{0.55\linewidth}
        \captionsetup{type=figure}
        \caption{Cosine similarity between the denoising directions, i.e., $\text{Cos}(\mathbf{x}_{t}-\hat{\mathbb{E}},\mathbf{x}_{t}-D)$, by empirical and neural denoisers. Here, $\mathbf{x}_{t}-\hat{\mathbb{E}}$ denotes the actual denoising direction induced by the update rule in \eqref{denoising}. Results are reported along the log-SNR axis~\citep{kingma2021variational}, where $\mathrm{SNR}=1/4t^{2}$, providing a unified noise-scale representation across diffusion formulations (e.g., EDM, DDPM, FM).}
  \label{fig:denoiser_comparison}
    \end{minipage}
\end{figure}

\subsection{Multi-Step Denoising is also a GDP}\label{sec:multi-step}

For discretized timesteps $t_i$ with increments $\Delta t_i$, the $i$-th empirical denoising step has $\mu_{t_i}=\frac{2C}{\beta|\mathcal{D}|}\sqrt{2\Delta t_i/t_i^3}$. By exact GDP composition, denoising from $t_m$ to $t_n$ is $\mu$-GDP with $\mu=(\sum_{i=n}^{m}\mu_{t_i}^2)^{1/2}$. The parameter $\mu$ has a direct testing interpretation: larger $\mu$ means it is easier to distinguish neighboring datasets from the generated output. Privacy loss is therefore far from uniform along the trajectory. Under the EDM schedule $\Delta t_i \approx c\, t_i^{1-1/\rho}$, so $\mu_{t_i} \propto t_i^{-1-1/(2\rho)}$, which is roughly $1/t_i$ for $\rho = 7$ (Figure~\ref{fig:placeholder}). The last low-noise steps therefore dominate the sum. This suggests a hybrid sampler: run the private empirical denoiser through the early and middle steps, then hand off to a public denoiser, whose steps are post-processing and add no privacy cost (Section~\ref{sec:experiment}).

\subsection{Bridging Empirical and Neural Denoisers}
\label{sec:bridging}

A central limitation is that real diffusion samplers use neural denoisers, while our formal accounting uses an empirical denoiser. Directly accounting for neural denoisers remains open because the sensitivity of a trained network to one training example is not tractable. We therefore use the empirical denoiser as an analyzable proxy and check where this proxy is informative.

First, Figure~\ref{fig:denoiser_comparison} shows that empirical and neural denoising directions are nearly aligned on CIFAR-10 in the earlier stages of the denoising process. Second, recent work on critical windows~\citep{li2024critical,biroli2024dynamical,sclocchi2024phase,georgiev2023journey,raya2023spontaneous} suggests that coarse semantic identity is determined early and then remains stable, so the formative part of the trajectory overlaps with the regime where the two denoisers are most similar. Together, these checks do not prove a neural-denoiser DP guarantee, but they justify using the empirical accountant to study where sampling leakage occurs.

Cosine similarity captures only the denoising \emph{direction}; the per-step magnitude gap $\|\widehat{\mathbb{E}}-D\|^2$ is reported in Appendix~\ref{app:mse}, where it peaks inside the mid-noise window, but as Table~\ref{tab:traj} shows, does not survive to the released sample. We therefore restrict the proxy claim to directional agreement over the regime our private block occupies.

\paragraph{Trajectory-level agreement.} The per-step divergence above bounds how far the two kernels can differ at a single step, but not how much of that difference survives to the released sample. We therefore substitute the empirical denoiser for the neural denoiser within one window at a time, with the initial noise and all stochastic increments held fixed, and measure the embedding distance ($1-\cos$) of the resulting sample to the all-neural reference trajectory. As a descriptive scale, we also keep the neural denoiser and re-draw only the noise increments inside that window. Table~\ref{tab:traj} reports both distances over 2048 paired trajectories, and Appendix~\ref{app:mse} adds absolute similarities, paired-bootstrap CIs, PSNR, SSIM, and label agreement \citep{efron1993introduction, wang2004image}. In the high- and mid-noise windows substitution moves the endpoint less than re-drawing that window's noise (mid-noise: $0.77\times$ CLIP, $0.83\times$ DINO, reference minus swap distance $+0.0228$ in CLIP, 95\% CI $[+0.0207, +0.0248]$). In the mid-noise window endpoint similarity stays high in CLIP ($0.923$) but falls in DINO ($0.483$), so coarse semantics are preserved while instance-level structure is not. In the low-noise window the ordering reverses (CLIP difference $-0.0441$, 95\% CI $[-0.0459, -0.0422]$), consistent with the $t^{-3}$ growth of the per-step divergence. The min-noise window is always handled by the public denoiser, while the low-noise window is private in our 65\% configuration (Section~\ref{subsec:privacy_critical_windows}).

Taken together, these measurements support a claim: through the mid-noise window, about the first half of generation, the empirical denoiser is a window-local semantic proxy for the neural denoiser. They do not establish a DP guarantee for a neural denoiser by itself, which remains open.

\begin{table}[t]
\centering\small
\caption{Endpoint distance ($1-\cos$) after substituting the empirical denoiser inside one window (swap) and after re-drawing only that window's noise with the neural denoiser kept (ref.). rel.\ = swap/ref.}
\label{tab:traj}
\begin{tabular}{llrrrrrr}
\toprule
& & \multicolumn{3}{c}{CLIP~\citep{radford2021learning}} & \multicolumn{3}{c}{DINO~\citep{oquab2024dinov2}} \\
\cmidrule(lr){3-5}\cmidrule(lr){6-8}
Window & $t$ range & swap & ref. & rel. & swap & ref. & rel. \\
\midrule
high-noise   & 19.15--7.47  & 0.0097 & 0.0917 & \textbf{0.11}$\times$ & 0.0717 & 0.5792 & \textbf{0.12}$\times$ \\
mid-noise & 6.09--1.98   & 0.0772 & 0.1000 & \textbf{0.77}$\times$ & 0.5175 & 0.6203 & \textbf{0.83}$\times$ \\
low-noise    & 1.54--0.38   & 0.1127 & 0.0687 & 1.64$\times$ & 0.6910 & 0.4431 & 1.56$\times$ \\
min-noise     & 0.14--0.018  & 0.2147 & 0.0241 & 8.90$\times$ & 0.8758 & 0.1376 & 6.37$\times$ \\
\bottomrule
\end{tabular}
\end{table}

\section{Privacy Analysis of Denoising Process with Subsampled Data}\label{sec:subsampling}
Section~\ref{sec:method} analyzes the empirical denoiser when every record in $\mathcal{D}$ contributes to every private denoising step. This full-batch accountant is tight, but it is expensive in terms of privacy: each generated output repeatedly touches every record, and privacy loss grows quickly when many outputs are released. 
Subsampling reduces this exposure because a record that is not selected contributes no privacy loss. 
Throughout, each generated output draws a minibatch $\mathcal{S}\subseteq\mathcal{D}$ of fixed size $B$ uniformly without replacement, so a fixed record is included with probability $q_{\mathcal{B}}=B/n$, and neighboring datasets differ by adding or removing one record. Because $|\mathcal{S}|=B$ is fixed, an added record that is selected takes the place of another record of $\mathcal{S}$, so within the minibatch the change is a replacement \citep{zhu2022optimal, lebeda2025avoiding}. The clipped minibatch average therefore has sensitivity $2C/B$, and each step is $\mu_t$-GDP with $\mu_t=\frac{2C}{\beta B}\sqrt{2\Delta t/t^3}$ conditional on inclusion.
The subsampled Gaussian mechanism is $\mathcal{M}_{\mathrm{SubGauss}}(\mathcal{D};\vx_t)=|\mathcal{S}|^{-1}\sum_{\vx_0\in\mathcal{S}}\bar{\mathbf{v}}(\vx_0;\vx_t)+\beta\sqrt{t^3/(2\Delta t)}\,\bm{\epsilon}$.

\begin{figure}[t]
    \centering
    \includegraphics[width=0.62\linewidth]{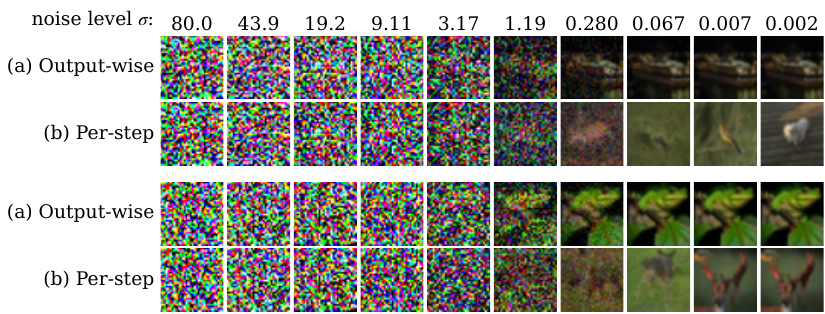}
    \caption{Output-wise vs. per-step subsampling, $B=25$, identical initial noise. Each row is one sample; columns show decreasing noise levels. Under output-wise subsampling (a) the subsample is reused, so each row progressively denoises the same emerging image. Under per-step subsampling (b) a fresh subsample is drawn each step, pulling the trajectory toward a different subset: semantic content changes between columns and never settles. This instability motivates block reuse.}
    \label{fig:sampling_comparison}
\end{figure}

The important design choice is how often we resample $\mathcal{S}$. Resampling at every denoising step for every output gives more independent amplification events, but it also changes the empirical denoising direction at every step. Small minibatches, which is better for privacy, inject extra stochastic drift into the reverse trajectory: the sample is repeatedly pulled toward different private subsets, and late low-noise steps are especially sensitive to this instability, as shown in Figure~\ref{fig:sampling_comparison}. This can harm sample quality even though the privacy accountant looks attractive. We therefore employ \emph{output-wise subsampling}: sample once for a generated output, reuse them through an empirical denoising block, then account for amplification once at the block level. This keeps the denoising direction stable while still limiting how often any record is accessed.

We state both accountants because they correspond to different mechanisms. Applying per-step amplification to output-wise subsampling would be incorrect: privacy amplification is tied to independent subsampling events, not to the number of denoising steps. If the same minibatch is reused over a block $\mathcal{B}$, the included record accumulates the full block-level Gaussian parameter $\bar{\mu}_{\mathcal{B}}=(\sum_{t\in\mathcal{B}}\mu_t^2)^{1/2}$ before amplification is applied. If a fresh minibatch is drawn at every step, each $\mu_t$ is amplified separately.

For full-batch Gaussian mechanisms, GDP composition is exact. With subsampling, each block's trade-off curve is a mixture rather than a Gaussian curve, so exact GDP composition no longer applies, and we compose numerically with a pessimistic PLD accountant (Section~\ref{subsec:pld}). The GDP central limit theorem~\citep{GDP} fits poorly here because the $\mu_t$ are highly heterogeneous, and its Berry-Esseen error is additive and too coarse for small $\delta$ (\tabref{edgeworth_vs_clt}). As a fast estimate, we also compute a degree-two Edgeworth approximation~\citep{zheng2020sharp}, which adds skewness and kurtosis corrections from the first four cumulants (\eqref{edgeworth}, Appendix~\ref{app:edgeworth_diagnostics}). It is an approximation and matches the certified PLD value to within 1\% for $\varepsilon \leq 2$ but understates it for larger $\varepsilon$, by up to 15\% in our runs (Appendix~\ref{app:pld_vs_ew}).

\begin{prop}[Output-wise subsampling]\label{prop:output_subsampling}
Consider a sampler that, for each of $N_{\mathrm{out}}$ generated outputs, performs private subsampling once and reuses them for an empirical denoising block $\mathcal{B}$; all later public denoising is post-processing. Let $q_{\mathcal{B}}$ be the block-level inclusion probability of a fixed record and $\bar{\mu}_{\mathcal{B}}=(\sum_{t\in\mathcal{B}}\mu_t^2)^{1/2}$. Define $h_{\mathcal{B}}=q_{\mathcal{B}}G_{\bar{\mu}_{\mathcal{B}}}+(1-q_{\mathcal{B}})\operatorname{Id}$ and
\begin{align}
    g_{\mathrm{out}}=\bigotimes_{j=1}^{N_{\mathrm{out}}}h_{\mathcal{B}},
    \qquad
    f_{\mathrm{out}}=\min\{g_{\mathrm{out}},g_{\mathrm{out}}^{-1}\}^{**}.
\end{align}
Then the joint release of the $N_{\mathrm{out}}$ generated outputs is $f_{\mathrm{out}}$-DP. Equivalently, for every $\varepsilon\ge0$ it is $(\varepsilon,\delta_{\mathrm{out}}(\varepsilon))$-DP with
\begin{align}
    \delta_{\mathrm{out}}(\varepsilon)
    =
    \max_{\mathrm{dir}\in\{P\to Q,Q\to P\}}
    \left[
    \Pr_Q(S_{\mathrm{out}}^{\mathrm{dir}}>\varepsilon)
    -
    e^\varepsilon
    \Pr_P(S_{\mathrm{out}}^{\mathrm{dir}}>\varepsilon)
    \right]_+,
\end{align}
where $S_{\mathrm{out}}=\sum_{j=1}^{N_{\mathrm{out}}}L_{\mathcal{B}}(Z_j)$,  $L_{\mathcal{B}}(z)=\log(1-q_{\mathcal{B}}+q_{\mathcal{B}}\exp(\bar{\mu}_{\mathcal{B}}z-\bar{\mu}_{\mathcal{B}}^2/2))$, and $\bigotimes$ indicates a tensor product. The probabilities are taken under the two neighboring-data hypotheses.
We evaluate $\delta_{\mathrm{out}}(\varepsilon)$ with a pessimistic PLD accountant, which gives the certified values we report, and use the Edgeworth expansion only as an estimate (Appendix~\ref{app:pld_vs_ew}).
\end{prop}

Algorithm~\ref{alg:row-subsampled} instantiates this: the subsample is drawn once per output (line 3) and reused across the block, so amplification is applied once to $\bar\mu_{\mathcal{B}}$ rather than per step.
In practice, each output enters the PLD accountant (Section~\ref{subsec:pld}) as one subsampled Gaussian mechanism with noise multiplier $1/\bar\mu_{\mathcal{B}}$ and sampling probability $q_{\mathcal{B}}$, whose two neighboring directions give exactly $h_{\mathcal{B}}$ and $h_{\mathcal{B}}^{-1}$. Composing $N_{\rm out}$ copies gives $\delta_{\rm out}(\varepsilon)$, and Appendix~\ref{app:pld_vs_ew} lists every step and setting. The proof is in Appendix~\ref{app:subsampling_proofs}, which also gives the accountant for per-step resampling (Proposition~\ref{prop:step_subsampling}). That accounting is valid only when the implementation truly resamples at every step, so it does not apply to our sampler.

Output-wise subsampling is not privacy-free: if $\sum_t\mu_t^2$ is large, the influence of an included record accumulates before amplification. This motivates our hybrid denoiser, which switches to a public neural denoiser after the empirical denoising block so that the remaining generation is post-processing. We use the Edgeworth approximation only when the composed privacy loss is in a stable finite-sample regime; Appendix~\ref{app:edgeworth_diagnostics} states these diagnostics.

\section{Related Work}\label{sec:related_work}
Differentially private training has been the most direct approach to reduce memorization in diffusion models. 
DP training of diffusion models builds on DP-SGD~\citep{DP_SGD,nvidia_DPDM,deepmind_DPDM,liu2024differentially} and protects the weights at a cost in sample quality.
Private sampling has mostly been studied for Bayesian posterior inference, where the samples are model parameters rather than data~\citep{10.5555/3104482.3104568,pmlr-v37-wangg15,pmlr-v89-li19a,DPHMC,pmlr-v202-zhang23aw}.
\citet{bertazzi2025differential} show that privacy loss in SGLD can concentrate in the final step, but their analysis needs strongly convex, Lipschitz objectives, which our empirical denoiser does not satisfy.

\paragraph{Sampling-time privacy for diffusion models.}
Closest to our setting, \citet{sehwagdifferentially} also analyze privacy at generation rather than training time, clip denoiser outputs, treat each step as a Gaussian mechanism, and delegate late denoising to a public model. Our analysis differs in three respects. First, their per-step sensitivity $C/k$ arises from a sample-and-aggregate ensemble of $k$ non-private models trained on disjoint shards, so the sensitivity floor is set by the ensemble size and private training is still required; we obtain sensitivity $2C/|\mathcal{D}|$ by clipping individual contributions to a single empirical average, with no training at all. Second, they adopt a per-sample private-prediction threat model and name composition across multiple released samples as open; we give an explicit joint-release guarantee over $N_{\mathrm{out}}$ outputs under standard DP, using subsampled $f$-DP with block-level amplification. Third, we localize privacy loss to the critical window and place the handoff at the measured crossover rather than at a fixed schedule fraction, validating the choice with a membership-inference proxy. \citet{wei2023inherent} study inherent privacy properties of \emph{discrete} denoising diffusion and likewise find that leakage concentrates at particular stages of generation. Their mechanism is the discrete transition kernel, while ours is the continuous Gaussian reverse SDE, where the noise already injected by the sampler serves as the mechanism noise. Our results are therefore a continuous-state counterpart to their staged view, and our claims are scoped accordingly.

\paragraph{Empirical denoisers.}
\citet{niedoba2024towards} characterize the empirical denoiser in the \emph{non-private} setting and analyze how trained neural denoisers depart from it. We build on this object directly: our contribution is the clipped, privatized variant and its sensitivity analysis, and their non-private characterization is complementary evidence for the proxy argument in Section~\ref{sec:bridging}.

\section{Experiments}\label{sec:experiment}
We study (i) full-batch versus subsampled accounting, (ii) where privacy loss accumulates relative to the critical window, and (iii) the privacy-utility trade-off of the hybrid sampler.

\paragraph{Implementation} 

  We implement our modified reverse denoising process in PyTorch \citep{paszke2019pytorch} and categorize our experiments into two regimes: (i) \textbf{full-batch}, where the entire dataset is used in each denoising step, and (ii) \textbf{output-wise subsampled SDEdit~\citep{meng2021sdedit}}, where a minibatch of private examples is subsampled once per generated output and reused for an empirical denoising block. We apply the private empirical denoising steps until time $t_0$, then add Gaussian noise with scale $(\sigma_{t_1}^2 - \sigma_{t_0}^2)^{1/2}$ to move the sample back to an SDEdit restart level $t_1>t_0$, and finally complete generation with a public neural denoiser from $t_1$ to $0$. Only the empirical subsampled block is privacy-accounted. Re-noising to a higher noise level lets the public neural denoiser correct artifacts introduced by the empirical step while incurring no additional privacy loss by post-processing.

We use the EDM schedule~\citep{karras2022elucidating} with $\sigma_{\min}=0.002$, $\sigma_{\max}=80$, $\rho=7$, $N=50$ steps for full-batch runs and $N=39$ for the hybrid pipeline, and the public normalization $\beta=1$ (Appendices~\ref{app:beta-ablation} and~\ref{app:hybrid_evaluation}).

\paragraph{Privacy Analysis} Given the noise schedule and sensitivity, we calculate the corresponding $\mu$-GDP for each diffusion step (\ref{eq:gdp_composition}). For the full-batch case, we derive the overall $\mu$-GDP using the $n$-fold composition of $\mu_i$-GDP mechanisms. For the output-wise subsampled SDEdit hybrid, we compute privacy using the certified PLD accounting library \citep{doroshenko2022connect} with grid spacing $h=10^{-5}$ and $\texttt{log\_mass\_truncation\_bound} = -50$, and the Edgeworth approximation from Section~\ref{sec:subsampling}, which approximates the per-output privacy-loss distribution under both neighbouring-data hypotheses using its first four cumulants. The subsequent re-noising and public neural denoising stages incur no additional privacy loss.

\begin{figure}[t]
    \centering
    \includegraphics[width=0.9\linewidth]{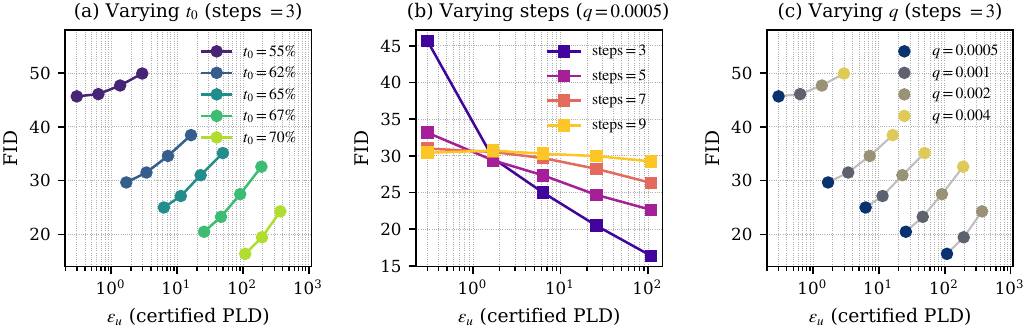}
    \caption{Privacy-utility trade-offs of the output-wise subsampled hybrid sampler on CIFAR-10.
    We report FID against the certified PLD upper bound $\varepsilon_u$ ($\delta=10^{-5}$, values from \tabref{hyperparam_search_rowsubsample}) while varying
    \textbf{(a)} the handoff time $t_0$,
    \textbf{(b)} the number of public re-noising steps from $t_0$ to $t_1$, and
    \textbf{(c)} the subsampling rate $q=B/n$.}
    \label{fig:hybrid_fid_eps}
\end{figure}

\paragraph{Datasets and Evaluation} On CIFAR-10~\citep{krizhevsky2009learning} ($n=50{,}000$) the public denoiser is trained on ImageNet $32\times32$~\citep{russakovsky2015imagenet}, and on CelebA $64\times64$~\citep{liu2015faceattributes} ($n=162{,}770$) it is trained on FFHQ~\citep{karras2019style}.
We use $\delta=10^{-5}\approx 1/n$ for CIFAR-10 and $10^{-6}$ for CelebA-64 to match prior DP generative models, and report the stricter $\delta=1/n^2$ in Appendix~\ref{app:eps-delta}.
Utility is FID against the training set with $50{,}000$ samples, reported as mean $\pm$ standard deviation over three generation seeds where runs are repeated.

\subsection{Privacy analysis of diffusion-based sampling with full training data}

When every record is used at every step, a single sample already costs $\varepsilon \approx 200$ when denoising down to $\sigma_{\min} = 0.1$, which is needed for a recognizable sample, and at $\varepsilon \approx 1$ ($\sigma_{\min} = 5$) the sample is indistinguishable from noise (Figure~\ref{fig:gdp_sigma_min}, Appendix~\ref{app:fullbatch}). Since each denoising step operates on the full dataset, the per-sample privacy loss is already high, and composing multiple samples results in a rapidly growing total privacy budget under GDP. Subsampling can alleviate this issue through privacy amplification, making it more suitable for large-scale sample generation.

\subsection{Hybrid denoising with output-wise subsampling}

\begin{table}[t]
\centering
\caption{Comparison of DP generative models on CIFAR-10 and CelebA-64. Our method performs a short private editing block using an output-wise subsampled empirical denoiser, then hands off to a publicly pretrained neural denoiser (ImageNet for CIFAR-10, FFHQ for CelebA-64) to complete sampling. FID is computed for $N_{out} = 50,000$ samples. Lower FID is better.}
\label{tab:dp_comparison}
\scalebox{0.65}
{\begin{tabular}{lcc@{\hskip 2em}lcc}
\toprule
\multicolumn{3}{c}{\textbf{CIFAR-10}} & \multicolumn{3}{c}{\textbf{CelebA-64}} \\
\cmidrule(r){1-3} \cmidrule(l){4-6}
Method & $\varepsilon$ & FID & Method & $\varepsilon$ & FID \\
\midrule
DP-Diffusion~\citep{deepmind_DPDM} & 10 & 9.8 & DP-LDM~\citep{liu2024differentially} & 10 & 14.3 \\
DP-LDM~\citep{liu2024differentially} & 10 & 8.4 & DP-MEPF~\citep{DP-MEPF} & 10 & 17.4 \\
DP-API~\citep{dp-api} & 0.67 & 7.87 & DP-DM~\citep{nvidia_DPDM} & 10 & 78.3 \\
DP-MEPF~\citep{DP-MEPF} & 10 & 29.1 & DP-Promise~\citep{dp_promise} & 10 & 25.3 \\
DP-DM~\citep{nvidia_DPDM} & 10 & 97.7 & & & \\
DP-Promise~\citep{dp_promise} & 10 & 17.9 & & & \\
\midrule
\textbf{Ours} & \makecell[c]{PLD $[11.14,\,11.38]$ \\[2pt] \footnotesize $\varepsilon_{ew} = 10$} & $27.1 \pm 0.15$
& \textbf{Ours} & \makecell[c]{PLD $[6.53,\,6.77]$  \\[2pt] \footnotesize $\varepsilon_{ew} = 5$} & $25.67 \pm 0.03$ \\
\bottomrule
\end{tabular}}
\end{table}

Figure~\ref{fig:hybrid_fid_eps} highlights three patterns. First, the empirical denoising stop timestep $t_0$ controls a clear privacy-utility tradeoff: stopping later at $t_0=70\%$ yields the best FID, while stopping earlier at $t_0=55\%$ yields the strongest privacy guarantee. Second, the best number of public re-noising steps depends on the privacy budget. With larger private blocks (65--70\%, certified $\varepsilon \geq 6$) three re-noising steps give the best FID, while with the smallest block (55\%, $\varepsilon \le 0.3$) more steps help (FID 45.7 with three steps and 30.4 with nine, $B = 25$). Re-noising steps are public, so they change utility but not $\varepsilon$. Third, decreasing the subsampling rate $q$ lowers $\varepsilon$, as expected from privacy amplification. In our setting it also improves FID, likely because smaller $q$ weakens the high-variance empirical denoiser before the public neural denoiser recovers global structure after re-noising.

To contextualize the effect of our privacy mechanisms, Table~\ref{tab:dp_comparison} compares the FID~\citep{heusel2017gans} of our hybrid sampler with DP generative models on CIFAR-10 and CelebA-64, and Appendix~\ref{app:hybrid_evaluation} reports all hybrid configurations.

\subsection{Privacy in relation to the critical windows}\label{subsec:privacy_critical_windows}

We extend our privacy analyses to quantify privacy levels based on where we stop relative to the critical window of the denoising process. Following \citet{georgiev2023journey}, we locate the critical window with a zero-shot CLIP classifier~\citep{radford2021learning} applied along the trajectory (Figure~\ref{fig:main}(a)).
The probability that the sample is classified as a ship is about 0.1 at $t = 0.27$, 0.3 at $t = 0.07$, and close to 1 at $t = 0.02$ (70\%, 80\%, and 88\% of the $N = 50$ steps).
Stopping the private block earlier gives a smaller $\mu$ but a noisier sample, and stopping later gives the reverse (Figure~\ref{fig:critical_window_full}).

We choose the handoff point so that, over the private block, the empirical denoiser remains a faithful stand-in for the neural denoiser.
The empirical denoiser is a window-local semantic proxy for the neural denoiser through the mid-noise window, which ends at $t = 1.98$, about half of the $N = 39$ steps (Table~\ref{tab:traj}), and directional agreement stays at $\cos \ge 0.98$ through about 55\% of generation (Figure~\ref{fig:denoiser_comparison}, Table~\ref{tab:mse}).
A 55\% handoff therefore keeps the private block essentially inside the proxy regime.
Our 65\% handoff ($t_0 \approx 0.52$) still comes before the CLIP-visible critical window (${\approx}\,70$--$88\%$), but its private block covers most of the low-noise window, where substitution moves the endpoint more than re-drawing that window's noise ($1.64\times$ CLIP, Table~\ref{tab:traj}).
The proxy interpretation is therefore weaker for the 65\% block, while the certified privacy guarantee holds for both, because it is a statement about the sampler we run.
Stopping earlier tightens the certified bound at a utility cost, and stopping later extends the private block into the low-noise steps that dominate the privacy cost (Appendix~\ref{app:hybrid_evaluation}).

\subsection{Privacy implications of neural and empirical denoisers in membership inference attacks}

We compare the membership-inference success rate (ASR) implied by our $\varepsilon$ for the subsampled empirical sampler (Appendix~\ref{sec:ASR}) with the ASR that the attack of \citet{duan2023are} achieves on a neural denoiser trained on all of CIFAR-10, stopping at the same $t$ and using the CIFAR-10 test set as non-members (\tabref{MIA}, Appendix~\ref{sec:ASR}).
The attack is weak at every $t$ (ASR at most 0.510) and stays below the bound for $q = 0.5$ at every $t$.
At $q \le 0.01$ it slightly exceeds the bound at $t = T/2$ (0.507 vs.\ at most 0.503), which is expected because the neural denoiser uses every record while the subsampled sampler includes each record with probability $q$.

\section{Conclusion and Discussion}
We give certified sampling-time privacy guarantees for continuous diffusion models by privatizing a clipped empirical denoiser. Framing each denoising step as a Gaussian mechanism, we account with GDP and $f$-DP, report certified $(\epsilon,\delta)$ bounds from a PLD accountant, and characterize where Edgeworth approximation is and is not reliable. Our results show that leakage is highly non-uniform across timesteps, concentrating in critical windows where semantic structure emerges, while subsampling significantly amplifies privacy guarantees. 
These findings suggest that privacy in diffusion models is not solely determined by training but can be shaped by sampling strategies such as early stopping and hybrid use of public denoisers. 

Our certificate covers the sampler we run. Samplers that query a neural denoiser trained on the private data at every step, and discrete diffusion, where the transition kernel would be the mechanism, remain open.
Our sampler could also serve as a DP data-augmentation mechanism $\mathcal{A}(\mathcal{D})$: a model trained only on $\mathcal{A}(\mathcal{D})$ and public data is $(\varepsilon,\delta)$-DP by post-processing, with no DP training algorithm. Since privacy loss concentrates at low noise, training on $\mathcal{A}(\mathcal{D})$ for $t \ge t_0$ and on public data for $t < t_0$ would spend the budget only on the formative regime.

\subsection*{AI use statement}

In this work, we used generative AI tools as writing, reviewing, and coding assistants.
Specifically, we used them to improve grammar and word choice, to draft and revise parts of the text, to get feedback on drafts, such as consistency checks of numbers, claims, and references and suggestions for additional diagnostics.
We also used them to search for and summarize related work, which we verified against the original sources, and to help write and debug code for the accounting, evaluation, and scripts, which we reviewed and tested before use. Additionally, we used AI to proofread our math equations and review our mathematical proofs.
We did not use generative AI tools to generate research ideas or synthetic data, or to produce any reported number.
Dataset cleaning, translation, and qualitative data analysis are not applicable to this work.
We have reviewed all AI-assisted work and take full responsibility for the final content of this paper, including text, claims, and artifacts produced with the aid of generative AI.

\subsection*{Ethics statement}

This work studies privacy leakage from diffusion sampling and uses only public benchmark datasets (CIFAR-10, CelebA, ImageNet, FFHQ).
Our guarantees cover the sampler in Algorithm~\ref{alg:row-subsampled} and should not be read as guarantees for other diffusion samplers (Appendix~\ref{app:impact_statement}).

\subsection*{Reproducibility statement}

The code is available in Appendix~\ref{app:code}, including the PLD accounting, Edgeworth, and FID scripts behind every table and figure.
Algorithm~\ref{alg:row-subsampled} specifies the sampler, Section~\ref{sec:method} and Appendix~\ref{app:beta-ablation} give the sensitivity and the public normalization, and Appendix~\ref{app:subsampling_proofs} gives full proofs.
Section~\ref{sec:experiment} and Appendix~\ref{app:pld_vs_ew} list the accountant settings (sampling model, adjacency, $q$, $N_{\mathrm{out}}$, $\delta$, grid spacing, truncation, and precision).
Appendix~\ref{app:hybrid_evaluation} lists every hyperparameter and reports all 95 CIFAR-10 and 112 CelebA-64 configurations (Tables~\ref{tab:hyperparam_search_rowsubsample} and~\ref{tab:celeba_hyperparam_main}) and the repeated-seed runs.

\subsection*{Acknowledgments}
We thank our anonymous reviewers for their constructive feedback, which has helped significantly improve our manuscript. We thank the Digital Research Alliance of Canada (Compute Canada) for its computational resources and services. W. Cheng and M. Park were supported in part by the Natural Sciences and Engineering Research Council of Canada (NSERC) and the Canada CIFAR AI Chairs program.

\newpage
\bibliography{main}
\bibliographystyle{iclr2027_conference}

\appendix
\section{Code Availability}\label{app:code}

The public code is available at \url{https://github.com/whalebeavercat/DPDiffusionSampling/}.

\section{Lipschitz-based sensitivity analysis for neural denoisers}\label{app:Lipschitz}

One might wonder whether considering Lipschitz-based sensitivity analysis for neural denoisers would help validate the empirical denoiser bounds. We describe our thoughts on this below. 

 Let $\mathcal L : X^n \to \Theta$ be the score network learning algorithm which maps a dataset $S \in X^n$ to parameters $\theta \in \Theta$. Let us define the sensitivity of the score network as $\Delta_s = \sup_{x \in X} \left( \sup_{S \in X^n, i \in [n]} \|{s_{\mathcal L(S)}(x) - s_{\mathcal L(S \setminus \{i\})}(x)}\|_2\right)$. If we could show that $\Delta_s$ is bounded, then the update step in sampling can be made differentially private with appropriate noise scales as follows: $\tilde x_t \leftarrow \tilde x_{t-1} + \frac{\sigma_i}{2} {s}_\theta(\tilde x_{t-1}, \sigma_i) + \mathcal N(0, \sigma_t^2 \Delta_s^2 I)$. The question is how to bound the sensitivity. Does the Lipschitz network help here? 

Suppose the score network is a 1-Lipschitz network, satisfying $\|s_{\mathcal L(S)}(x) - s_{\mathcal L(S)}(x')\|_2 \leq \|x-x'\|_2$ and $\|s_{\mathcal L(S \setminus \{i\})}(x)-s_{\mathcal L(S \setminus \{i\})}(x')\|_2 \leq \|x-x'\|_2$. Using this we can bound the sensitivity, $\|s_{\mathcal{L(S)}}(x)-s_{\mathcal L(S \setminus \{i\})}(x)\|_2 \leq \|s_{\mathcal{L(S)}}(x)-s_{\mathcal{L(S)}}(x')\|_2 + \|s_{\mathcal{L(S)}}(x')- s_{\mathcal L(S \setminus \{i\})}(x)\|_2 \leq \|x-x'\|_2 + \|s_{\mathcal{L(S)}}(x')- s_{\mathcal L(S \setminus \{i\})}(x)\|_2$. So here even if $\|s_{\mathcal{L(S)}}(x')- s_{\mathcal L(S \setminus \{i\})}(x)\|_2$ is very small, we have $\max_{x, x'}\|x-x'\|_2$ on our upperbound. 
 
Without having any knowledge other than all $x, x'$ are pre-processed such that any of them is norm-bounded by $L$, the upperbound to this is $\max_{x, x'}\|x-x'\|_2 \leq \max_x \|x\|_2 \leq 2L$. In this case, the amount of noise we have to add in each denoising step is on the order of the entire input domain, and such an amount of noise will destroy the signal from the score net completely.  Hence, this approach is not usable in practice. 

\section{Proofs for Subsampling Propositions}\label{app:subsampling_proofs}

We prove Proposition~\ref{prop:output_subsampling} and Proposition~\ref{prop:step_subsampling} (per-step subsampling, stated below) under the sampling model stated in Section~\ref{sec:subsampling}. The proofs are directional first, comparing two neighboring datasets $\mathcal{D}$ and $\mathcal{D}'$ that differ by adding or removing one record, and then symmetrized by taking both neighboring-data directions.

\subsection{Proof of Proposition~\ref{prop:output_subsampling}}
\label{app:output_subsampling_proof}

Fix one generated output and an empirical denoising block $\mathcal{B}$, and let $\mathcal{D}'=\mathcal{D}\cup\{x\}$, so that $q_{\mathcal{B}}=B/n$ with $n=|\mathcal{D}'|$. A uniform $B$-subset of $\mathcal{D}'$ excludes $x$ with probability $1-q_{\mathcal{B}}$, and it is then a uniform $B$-subset of $\mathcal{D}$, so the two neighboring datasets induce the same output distribution and the trade-off curve is $\operatorname{Id}$. With probability $q_{\mathcal{B}}$, it is $S'\cup\{x\}$, where $S'$ is a uniform $(B-1)$-subset of $\mathcal{D}$. We pair it with $S'\cup\{y\}$, where $y$ is uniform in $\mathcal{D}\setminus S'$, which is a uniform $B$-subset of $\mathcal{D}$. These two minibatches differ by one replacement, so each step has parameter $\mu_t$ (Section~\ref{sec:subsampling}), and by GDP composition the block is $\bar\mu_{\mathcal{B}}$-GDP for every such pair, where
\begin{align}
    \bar\mu_{\mathcal{B}}=\Big(\sum_{t\in\mathcal{B}}\mu_t^2\Big)^{1/2}.
\end{align}
The hockey-stick divergence is jointly convex, so this bound survives averaging over $(S',y)$, and the canonical binary experiment for the included-record event is $P_{\mathcal{B}}=\mathcal{N}(0,1)$ versus $\mathcal{N}(\bar\mu_{\mathcal{B}},1)$, with trade-off curve $G_{\bar\mu_{\mathcal{B}}}$. Thus the one-output directional trade-off curve is
\begin{align}
    h_{\mathcal{B}}
    =
    q_{\mathcal{B}}G_{\bar{\mu}_{\mathcal{B}}}
    +
    (1-q_{\mathcal{B}})\operatorname{Id}.
\end{align}

The $N_{\mathrm{out}}$ generated outputs use independent row-subsampling events, so $f$-DP composition gives the directional composed trade-off
\begin{align}
    g_{\mathrm{out}}
    =
    \bigotimes_{j=1}^{N_{\mathrm{out}}} h_{\mathcal{B}}.
\end{align}
Since differential privacy must hold for both neighboring-data directions, we symmetrize the directional curve and take the lower convex envelope, yielding
\begin{align}
    f_{\mathrm{out}}
    =
    \min\{g_{\mathrm{out}},g_{\mathrm{out}}^{-1}\}^{**}.
\end{align}
Therefore the joint release is $f_{\mathrm{out}}$-DP. Any public neural denoising or deterministic post-processing after the private block is post-processing of the same private release and cannot weaken the guarantee.

It remains to derive the $(\varepsilon,\delta)$ profile. In the above directional experiment, the privacy-loss random variable for one output can be written as
\begin{align}
    L_{\mathcal{B}}(z)
    =
    \log\left(
    1-q_{\mathcal{B}}
    +
    q_{\mathcal{B}}
    \exp\left(
    \bar{\mu}_{\mathcal{B}}z-\frac{\bar{\mu}_{\mathcal{B}}^2}{2}
    \right)
    \right),
\end{align}
and the composed privacy loss is $S_{\mathrm{out}}=\sum_{j=1}^{N_{\mathrm{out}}}L_{\mathcal{B}}(Z_j)$. The hockey-stick divergence characterization of approximate DP gives
\begin{align}
    \delta_{\mathrm{out}}^{P\to Q}(\varepsilon)
    =
    \Pr_Q(S_{\mathrm{out}}>\varepsilon)
    -
    e^\varepsilon\Pr_P(S_{\mathrm{out}}>\varepsilon).
\end{align}
Applying the same argument in the reverse neighboring direction and taking the positive part gives the stated $\delta_{\mathrm{out}}(\varepsilon)$.

\subsection{Per-step subsampling}\label{app:step_subsampling}

\begin{prop}[Per-step subsampling]\label{prop:step_subsampling}
Consider instead a sampler that draws an independent minibatch of fixed size $B_t$ at every private denoising step. For step $t$, let $q_t = B_t/n$ be the inclusion probability and $\mu_t$ the Gaussian parameter conditional on inclusion, and set $h_t=q_tG_{\mu_t}+(1-q_t)\operatorname{Id}$. For $N_{\mathrm{out}}$ outputs and $T$ private steps, define
\begin{align}
    g_{\mathrm{step}}
    =
    \bigotimes_{j=1}^{N_{\mathrm{out}}}
    \bigotimes_{t=1}^{T}h_t,
    \qquad
    f_{\mathrm{step}}
    =
    \min\{g_{\mathrm{step}},g_{\mathrm{step}}^{-1}\}^{**}.
\end{align}
Then the joint release is $f_{\mathrm{step}}$-DP. Hence, for every $\varepsilon\ge0$ it is $(\varepsilon,\delta_{\mathrm{step}}(\varepsilon))$-DP, where $\delta_{\mathrm{step}}(\varepsilon)=\max_{\mathrm{dir}}[\Pr_Q(S_{\mathrm{step}}^{\mathrm{dir}}>\varepsilon)-e^\varepsilon\Pr_P(S_{\mathrm{step}}^{\mathrm{dir}}>\varepsilon)]_+$, $S_{\mathrm{step}}=\sum_{j=1}^{N_{\mathrm{out}}}\sum_{t=1}^{T}L_t(Z_{j,t})$, and $L_t(z)=\log(1-q_t+q_t\exp(\mu_tz-\mu_t^2/2))$.
\end{prop}

The two propositions describe different privacy-loss distributions. In general, $S_{\mathrm{step}}=\sum_{j,t}L_t(Z_{j,t})\neq\sum_jL_{\mathcal{B}}(Z_j)=S_{\mathrm{out}}$, so $f_{\mathrm{step}}\neq f_{\mathrm{out}}$. Per-step amplification is valid only when the implementation truly resamples at every step. For our output-wise subsampled sampler, Proposition~\ref{prop:output_subsampling} is the correct accountant: amplification is applied once to the whole empirical block.

\subsection{Proof of Proposition~\ref{prop:step_subsampling}}

For per-step subsampling, the differing record is included at step $t$ with probability $q_t$, independently across steps. As in Appendix~C.1, each minibatch of $\mathcal{D}'$ containing $x$ is paired with a minibatch of $\mathcal{D}$ that differs from it in one record, now with $B_t$, $q_t$ and $\mu_t$ in place of $B$, $q_{\mathcal{B}}$ and $\bar\mu_{\mathcal{B}}$. Conditional on inclusion at step $t$, the step is a Gaussian mechanism with GDP parameter $\mu_t$, so the canonical included-record trade-off curve is $G_{\mu_t}$. If the record is not included, the two neighboring datasets induce the same distribution for that step and the trade-off curve is $\operatorname{Id}$. Therefore the one-step directional trade-off curve is
\begin{align}
    h_t
    =
    q_tG_{\mu_t}
    +
    (1-q_t)\operatorname{Id}.
\end{align}

Because minibatches are redrawn independently across timesteps and generated outputs, the directional trade-off curve for $N_{\mathrm{out}}$ outputs and $T$ private steps is the tensor-product composition
\begin{align}
    g_{\mathrm{step}}
    =
    \bigotimes_{j=1}^{N_{\mathrm{out}}}
    \bigotimes_{t=1}^{T} h_t.
\end{align}
Symmetrizing over the two neighboring-data directions and convexifying gives
\begin{align}
    f_{\mathrm{step}}
    =
    \min\{g_{\mathrm{step}},g_{\mathrm{step}}^{-1}\}^{**},
\end{align}
so the joint release is $f_{\mathrm{step}}$-DP.

The corresponding likelihood-ratio representation is obtained by summing the independent per-step privacy losses. For one step,
\begin{align}
    L_t(z)
    =
    \log\left(
    1-q_t
    +
    q_t\exp\left(\mu_tz-\frac{\mu_t^2}{2}\right)
    \right),
\end{align}
and hence
\begin{align}
    S_{\mathrm{step}}
    =
    \sum_{j=1}^{N_{\mathrm{out}}}\sum_{t=1}^{T}L_t(Z_{j,t}).
\end{align}
Using the same hockey-stick divergence conversion as above gives
\begin{align}
    \delta_{\mathrm{step}}(\varepsilon)
    =
    \max_{\mathrm{dir}}
    \left[
    \Pr_Q(S_{\mathrm{step}}^{\mathrm{dir}}>\varepsilon)
    -
    e^\varepsilon\Pr_P(S_{\mathrm{step}}^{\mathrm{dir}}>\varepsilon)
    \right]_+,
\end{align}
which is the claimed $(\varepsilon,\delta)$ guarantee.

\section{Diagnostics for Edgeworth Approximation}\label{app:edgeworth_diagnostics}

The standard Edgeworth approximation for subsampled Gaussian mechanisms is often stated for the setting where a fresh subsample is drawn at every private step. Our output-wise subsampling mechanism uses the same mixture privacy-loss model, but after grouping timesteps by independent subsampling events. Thus each ``step'' in the diagnostic below should be read as a block $\mathcal{B}$ with sampling probability $q_{\mathcal{B}}$ and Gaussian parameter
\begin{align}
    \bar{\mu}_{\mathcal{B}}
    =
    \left(\sum_{t\in\mathcal{B}}\mu_t^2\right)^{1/2}.
\end{align}
For a block, recall the privacy-loss variable
\begin{align}
    L_{\mathcal{B}}(z)
    =
    \log\left(
    1-q_{\mathcal{B}}
    +
    q_{\mathcal{B}}
    \exp\left(
    \bar{\mu}_{\mathcal{B}}z
    -
    \frac{\bar{\mu}_{\mathcal{B}}^2}{2}
    \right)
    \right),
\end{align}
evaluated under both
$P_{\mathcal{B}}=\Nrm(0,1)$ and
$Q_{\mathcal{B}}=(1-q_{\mathcal{B}})\Nrm(0,1)+q_{\mathcal{B}}\Nrm(\bar{\mu}_{\mathcal{B}},1)$.
Let $\kappa^D_{r,\mathcal{B}}$ denote the $r$-th cumulant of $L_{\mathcal{B}}$ under $D\in\{P,Q\}$. Across generated samples and independent blocks, cumulants add:
\begin{align}
    M_D=\sum_{\mathcal{B}}\kappa^D_{1,\mathcal{B}},
    \qquad
    V_D=\sum_{\mathcal{B}}\kappa^D_{2,\mathcal{B}},
    \qquad
    K_{3,D}=\sum_{\mathcal{B}}\kappa^D_{3,\mathcal{B}},
    \qquad
    K_{4,D}=\sum_{\mathcal{B}}\kappa^D_{4,\mathcal{B}}.
\end{align}

\paragraph{Edgeworth approximation.}
Let $S$ denote a composed privacy-loss variable with mean $M$, variance $V$, third cumulant $K_3$, and fourth cumulant $K_4$ (that is, $M_D$, $V_D$, $K_{3,D}$, $K_{4,D}$ under $D\in\{P,Q\}$). The degree-two Edgeworth approximation to the CDF of $(S-M)/\sqrt{V}$ is
\begin{align}\label{eq:edgeworth}
    F_{\mathrm{EW}}(z)
    =
    \Phi(z)
    -
    \varphi(z)
    \left[
    \frac{K_3}{6V^{3/2}}(z^2-1)
    +
    \frac{K_4}{24V^2}(z^3-3z)
    +
    \frac{K_3^2}{72V^3}(z^5-10z^3+15z)
    \right],
\end{align}
where $\Phi$ and $\varphi$ are the standard normal CDF and density. We apply this approximation separately under the two neighboring-data hypotheses, since the two privacy-loss CDFs need not match.

\paragraph{What the cumulants represent.}
For a random variable $X$, the cumulant generating function is
\begin{align}
    K_X(s)=\log \mathbb{E}[e^{sX}],
\end{align}
and the $r$-th cumulant is the $r$-th derivative of $K_X$ at $s=0$. In our accountant, $X=L_{\mathcal{B}}$ under either $P$ or $Q$. The first four cumulants have the familiar interpretations
\begin{align}
    \kappa_1 &= \mathbb{E}[X], \\
    \kappa_2 &= \mathrm{Var}(X), \\
    \kappa_3 &= \mathbb{E}[(X-\mathbb{E}X)^3], \\
    \kappa_4 &= \mathbb{E}[(X-\mathbb{E}X)^4]-3\mathrm{Var}(X)^2.
\end{align}
Thus $\kappa_1$ and $\kappa_2$ give the mean and variance of the privacy loss, while $\kappa_3$ and $\kappa_4$ quantify the leading skewness and kurtosis corrections to the Gaussian approximation.

\paragraph{No lower bound on the sampling rate alone.}
A small sampling probability $q_{\mathcal{B}}$ does not by itself invalidate the Edgeworth approximation. For fixed finite $\bar{\mu}_{\mathcal{B}}$, the variance and higher cumulants shrink with powers of $q_{\mathcal{B}}$, and these powers largely cancel after standardization. The relevant question is not whether $q_{\mathcal{B}}$ is large, but whether the composed privacy-loss sum has enough effective variance and is not dominated by a small number of blocks.

\paragraph{Variance and dominance checks.}
First, $V_P$ and $V_Q$ should be positive and numerically stable. If either variance is extremely small, the mechanism may be close to perfect privacy, but the standardized Edgeworth CDF and inverse-CDF calculations become ill-conditioned. Second, no block should dominate the variance:
\begin{align}
    \max_{\mathcal{B}}
    \frac{\kappa^P_{2,\mathcal{B}}}{V_P}
    \ll 1,
    \qquad
    \max_{\mathcal{B}}
    \frac{\kappa^Q_{2,\mathcal{B}}}{V_Q}
    \ll 1.
\end{align}
This condition is particularly important for diffusion schedules, where late low-noise timesteps can dominate $\sum_t\mu_t^2$ if they are grouped into a large private block.

\paragraph{Standardized cumulants.}
The Edgeworth correction should be small at the quantiles used for the reported $(\varepsilon,\delta)$ guarantee. We monitor the standardized third and fourth cumulants
\begin{align}
    \rho_{3,P}=\frac{|K_{3,P}|}{V_P^{3/2}},
    \qquad
    \rho_{4,P}=\frac{|K_{4,P}|}{V_P^2},
    \qquad
    \rho_{3,Q}=\frac{|K_{3,Q}|}{V_Q^{3/2}},
    \qquad
    \rho_{4,Q}=\frac{|K_{4,Q}|}{V_Q^2}.
\end{align}
Small values near the center of the distribution are not sufficient by themselves: degree-two Edgeworth terms multiply these quantities by polynomials in the standardized quantile $z$, so far-tail evaluations can still be unreliable.

\paragraph{Tail and CDF checks.}
Very small $\delta$ values may require evaluating the approximation close to the boundary of the CDF, where finite-sample Edgeworth approximations are less stable. We therefore check that the approximated CDFs are monotone, remain in $[0,1]$ before substantial clipping or projection, and produce a nonnegative $\delta(\varepsilon)$ curve in the reported range. We compute both neighboring-data directions and use the larger value. The Edgeworth value is never a certified bound. These diagnostics only decide whether the estimate is stable enough to show next to the certified PLD value, and configurations that fail them are marked invalid (\tabref{celeba_hyperparam_main}).

\paragraph{Practical implications of output-wise subsampling.}
Suppose the dataset has size $n=|\mathcal{D}|$ and each generated sample draws a minibatch of size $B$ under the sampling model of Section~\ref{sec:subsampling}, reused for an empirical denoising block. For a fixed training record, the block-level inclusion probability is
\begin{align}
    q_{\mathcal{B}}
    =
    B/n.
\end{align}
For our CIFAR-10 setting with $n=50{,}000$ and $B=25$, this gives $q_{\mathcal{B}}=5\times 10^{-4}$. If we release $N_{\mathrm{out}}=50{,}000$ generated samples, then a fixed record is included in about
\begin{align}
    N_{\mathrm{out}}q_{\mathcal{B}}= 25
\end{align}
generated trajectories on average. The privacy-relevant point is that each inclusion can affect the entire empirical block. Therefore the output-wise subsampled accountant is not
\begin{align}
    N_{\mathrm{out}}T
    \text{ subsampled Gaussian mechanisms with parameters }
    (q_t,\mu_t),
\end{align}
which would correspond to independently resampling at every denoising step. It is instead
\begin{align}
    N_{\mathrm{out}}
    \text{ subsampled block mechanisms with parameters }
    \left(q_{\mathcal{B}},
    \left(\sum_{t\in\mathcal{B}}\mu_t^2\right)^{1/2}\right).
\end{align}
This is the practical reason to subsample per output once and keep the empirical block short: output-wise subsampling sharply limits how often any record is accessed, but once a record is selected its influence accumulates across all empirical denoising steps in that block.

\begin{algorithm}[t]
\caption{Output-wise subsampled empirical denoising sampler}
\label{alg:row-subsampled}
\begin{algorithmic}[1]
\Require $\mathcal{D}$ with $n=|\mathcal{D}|$, subsample size $B$, clip norm $C$, EDM schedule with switch time $t_0$ and restart level $t_1>t_0$, public denoiser $D_{\mathrm{pub}}$ (data disjoint from $\mathcal{D}$), outputs $N_{\mathrm{out}}$.
\Ensure $N_{\mathrm{out}}$ samples with an $f_{\mathrm{out}}$-DP guarantee on $\mathcal{D}$.
\State $q_{\mathcal{B}} \gets B/n$;\quad
       $\mathcal{B}\gets\{\,t : t_0 < t \le t_{\max}\,\}$
  \Comment{block inclusion prob.; private block}
\For{$j=1$ \textbf{to} $N_{\mathrm{out}}$}
  \State draw $S_j\subseteq\mathcal{D}$, $|S_j|=B$, and \textbf{reuse for all of $\mathcal{B}$};\quad
         $x \gets \sigma_{\max}\epsilon$
    \Comment{subsample \emph{once} per output}
  \For{$t\in\mathcal{B}$ in decreasing order}
    \State $\hat{\mathbb{E}}\gets \tfrac{1}{\beta |S_j|}\sum_{x_0\in S_j}\bar v(x_0;x)$,\quad
           $\bar v = v/\max(1,\|v\|/C)$, \quad $v = x_0\,p_t(x\mid x_0)$
      \Comment{clipped direction}
    \State $x \gets \big(1-\tfrac{2\Delta t}{t}\big)x + \tfrac{2\Delta t}{t}\hat{\mathbb{E}} + \sqrt{2t\Delta t}\,\epsilon$
      \Comment{Gaussian mechanism, $\mu_t$ by \eqref{norm_clipped}}
  \EndFor
  \State $x \gets x + (\sigma_{t_1}^2-\sigma_{t_0}^2)^{1/2}\epsilon$;\quad
         $x^{(j)} \gets D_{\mathrm{pub}}$ integrated from $t_1$ to $0$
    \Comment{re-noise; public refinement (post-processing)}
\EndFor
\State $\bar\mu_{\mathcal{B}}\gets(\sum_{t\in\mathcal{B}}\mu_t^2)^{1/2}$; amplify \emph{once} by $q_{\mathcal{B}}$ and compose over $N_{\mathrm{out}}$ outputs (Prop.~\ref{prop:output_subsampling}) to obtain $f_{\mathrm{out}}$.
\State \Return $\{x^{(j)}\}_{j=1}^{N_{\mathrm{out}}}$ with certificate $f_{\mathrm{out}}$
\end{algorithmic}
\end{algorithm}

\section{Effect of the public normalization constant \texorpdfstring{$\beta$}{beta}}
\label{app:beta-ablation}

For EDM, the forward transition kernel is $p(x_t \mid x_0) = \mathcal{N}(x_t; x_0, t^2 I)$, so the marginal at each time is $\widehat p_t(x_t) = \frac{1}{|\mathcal{D}|}\sum_{x_0 \in \mathcal{D}} p(x_t \mid x_0)$. The exact empirical posterior mean divides the weighted average by this data-dependent normalizer:
\begin{equation}
  \widehat E[x_0 \mid x_t]
  = \frac{1}{\widehat p_t(x_t)\,|\mathcal{D}|}
    \sum_{x_0 \in \mathcal{D}} x_0\, p(x_t \mid x_0).
\end{equation}
Privately estimating and using this normalizer would require accounting for an additional data-dependent statistic. Our implemented mechanism instead fixes a public constant (hyperparameter) $\beta$ and uses
\begin{equation}
  \widehat E_\beta[x_0 \mid x_t]
  = \frac{1}{\beta\,|\mathcal{D}|}
    \sum_{x_0 \in \mathcal{D}} x_0\, p(x_t \mid x_0).
\end{equation}
Consequently, $\beta = 1$ defines a deliberately \emph{unnormalized empirical mechanism}; it is not an estimate that the density $\widehat p_t(x_t)$ numerically equals one. With public $\beta$, this remains a well-defined randomized mechanism with a rescaled empirical drift, and the privacy guarantee applies exactly to the implemented mechanism. Its privacy and utility both depend on the public scaling through
\begin{equation}
  \mu_t(\beta) = \frac{2C}{\beta\,|\mathcal{D}|}\sqrt{\frac{2\Delta t}{t^3}},
\end{equation}
where larger $\beta$ reduces both the private drift and $\mu_t$, while smaller $\beta$ strengthens both.

Table~\ref{tab:beta-ablation} reports $\beta$ directly (the experimental code uses the inverse, $1/\beta$). The setup is CIFAR-10, the $65\%$ empirical block, batch size $50$, three-step SDEdit restart, $\delta = 10^{-5}$, and $N_{\mathrm{out}} = 50{,}000$.

\begin{table}[h]
  \centering
  \caption{Effect of the public normalization $\beta$ on privacy and utility.}
  \label{tab:beta-ablation}
  \begin{tabular}{rrrr}
    \toprule
    $\beta$ & Edgeworth estimate & Certified PLD upper bound & FID \\
    \midrule
    4 & 0.438 & 0.438 & 132.89 \\
    2 & 1.218 & 1.217 & 62.81 \\
    1 & 10.777 & 11.381 & 27.13 \\
    0.5 & 256.624 & 259.761 & 53.54 \\
    0.25 & \multicolumn{1}{c}{unstable} & \multicolumn{1}{c}{unstable} & 81.43 \\
    \bottomrule
  \end{tabular}
\end{table}

This ablation shows that the public normalization is a genuine privacy--utility parameter rather than a privacy-neutral approximation: a weak empirical drift gives strong privacy but poor utility, whereas an overly strong drift worsens the certified privacy value and destabilizes generation.

\section{Per-step magnitude gap between empirical and neural denoisers}
\label{app:mse}

The relevant diagnostic in the main text is Figure~\ref{fig:denoiser_comparison} (directional cosine
similarity), which we supplement here with the mean-squared error (MSE) between the
neural and empirical denoiser outputs along the trajectory.

\begin{table}[h]
\centering
\caption{Directional cosine similarity (Fig.~\ref{fig:denoiser_comparison}) and per-step MSE
$\|\widehat{\mathbb{E}}-D\|^2$ between empirical and neural denoiser outputs, along the
sampling trajectory. Rows are labelled by the denoiser active at that noise level under
the hybrid sampler, and \textbf{switch} marks the empirical$\to$public handoff for the
$55/65/70\%$ configurations. Rows use the $N = 50$ EDM grid (percent = step$/N$). On the $N = 39$ grid of the hybrid runs these handoffs fall near $t = 1.5$, $0.52$, and $0.28$. MSE is mean $\pm$ standard error over 3 seeds.}
\label{tab:mse}
\begin{tabular}{rrrrrl}
\toprule
$t$ & \% gen. & log SNR & $\cos(x_t-\widehat{\mathbb{E}},\,x_t-D)$ & MSE & denoiser \\
\midrule
$80.0$   & $0\%$  & $-10.2$ & $1.0000$ & $3.24\pm0.03$    & emp. \\
$10.2$   & $32\%$ & $-6.0$  & $1.0000$ & $1.84\pm0.07$    & emp. \\
$3.29$   & $46\%$ & $-3.8$  & $0.9965$ & $14.63\pm0.20$   & emp. \\
$1.29$   & $55\%$ & $-1.9$  & $0.9895$ & $10.13\pm0.09$   & \textbf{switch 55\%} \\
$0.55$   & $65\%$ & $-0.2$  & $0.9766$ & $6.40\pm0.05$    & \textbf{switch 65\%} \\
$0.27$   & $70\%$ & $+1.3$  & $0.9568$ & $4.23\pm0.04$    & \textbf{switch 70\%} \\
$0.09$   & $78\%$ & $+3.4$  & $0.9011$ & $2.13\pm0.02$    & pub. \\
$0.02$   & $88\%$ & $+6.7$  & $0.7517$ & $0.62\pm0.004$   & pub. \\
$0.002$  & $98\%$ & $+11.0$ & $0.3901$ & $0.10\pm0.0003$  & pub. \\
\bottomrule
\end{tabular}
\end{table}

Cosine and MSE measure different quantities: MSE is $\|\widehat{\mathbb{E}}-D\|^2$, while
Figure~\ref{fig:denoiser_comparison} reports the cosine between the residual directions
$r_E=x_t-\widehat{\mathbb{E}}$ and $r_N=x_t-D$, related by
$\|\widehat{\mathbb{E}}-D\|^2=\|r_E\|^2+\|r_N\|^2-2\|r_E\|\|r_N\|\cos(r_E,r_N)$.
The cosine can be near one while the magnitudes differ (large MSE), and at low noise both
residual norms shrink, making the cosine numerically sensitive. Cosine alone therefore
cannot establish closeness; we report MSE alongside it and narrow the proxy claim to
directional agreement.

Absolute MSE is not comparable across rows: both residual norms shrink as $t\to0$, so the
small values at low noise reflect this scale collapse rather than closer agreement, and
the only cross-row claim we make is the directional one.

The magnitude gap is largest in the mid-noise window (the $46$--$55\%$ rows), yet
Table~\ref{tab:traj} shows that the difference surviving to the released sample
there stays below the noise-resampling reference: the private block sets the denoising
\emph{direction}, and the subsequent re-noising and public refinement
(Sec.~\ref{sec:experiment}) repair magnitude at no additional privacy cost.
Table~\ref{tab:traj_full} gives the absolute endpoint agreement behind Table~\ref{tab:traj}, and Table~\ref{tab:traj_image} gives image-space agreement for both the substituted and the noise-resampled paths.

\begin{table}[h]
\centering
\caption{Endpoint agreement between the all-neural reference and the path with the empirical denoiser substituted in one window (2048 paired trajectories, $N = 39$). CLIP and DINO are cosine similarities ($1 -$ the swap distance of Table~\ref{tab:traj}), with 95\% bootstrap CIs in brackets. $\Delta$CLIP is noise-control distance minus swap distance, with 95\% paired-bootstrap CIs. Image-space metrics are in Table~\ref{tab:traj_image}.}
\label{tab:traj_full}
\small
\setlength{\tabcolsep}{4pt}
\begin{tabular}{lllll}
\toprule
Window & $t$ range & CLIP & DINO & $\Delta$CLIP [95\% CI] \\
\midrule
high-noise & 19.15--7.47  & 0.990 [0.989, 0.991] & 0.928 [0.922, 0.934] & $+0.0821$ $[+0.0801, +0.0840]$ \\
mid-noise  & 6.09--1.98   & 0.923 [0.921, 0.925] & 0.483 [0.474, 0.491] & $+0.0228$ $[+0.0207, +0.0248]$ \\
low-noise  & 1.54--0.38   & 0.887 [0.886, 0.889] & 0.309 [0.303, 0.315] & $-0.0441$ $[-0.0459, -0.0422]$ \\
min-noise  & 0.14--0.018  & 0.785 [0.783, 0.788] & 0.124 [0.119, 0.129] & $-0.1906$ $[-0.1930, -0.1881]$ \\
all steps  & 80--0.002    & 0.794 [0.792, 0.796] & 0.121 [0.117, 0.126] & -- \\
\bottomrule
\end{tabular}
\end{table}

\begin{table}[h]
\centering
\caption{Image-space agreement between the all-neural endpoint and each variant, from the same starting noise (2048 trajectories, $N = 39$, windows of 6 steps). \emph{swap}: empirical denoiser substituted inside the window. \emph{ref.}: neural denoiser kept and only that window's noise re-drawn. Higher means closer to the all-neural endpoint for all three metrics. Brackets are 95\% paired bootstrap CIs.}
\label{tab:traj_image}
\small
\setlength{\tabcolsep}{4pt}
\begin{tabular}{llccc}
\toprule
Window (steps) & & PSNR (dB) & SSIM & Label agreement \\
\midrule
high-noise (9--14) & swap & 33.19 [32.88, 33.50] & 0.981 [0.979, 0.983] & 0.867 [0.853, 0.882] \\
 & ref. & 11.77 [11.67, 11.86] & 0.634 [0.629, 0.639] & 0.412 [0.391, 0.433] \\
\midrule
mid-noise (15--20) & swap & 14.74 [14.63, 14.84] & 0.731 [0.726, 0.736] & 0.462 [0.440, 0.483] \\
 & ref. & 10.40 [10.36, 10.45] & 0.466 [0.463, 0.469] & 0.390 [0.368, 0.411] \\
\midrule
low-noise (21--26) & swap & 12.23 [12.19, 12.27] & 0.412 [0.410, 0.415] & 0.255 [0.236, 0.274] \\
 & ref. & 11.98 [11.94, 12.03] & 0.415 [0.412, 0.419] & 0.603 [0.582, 0.624] \\
\midrule
min-noise (29--34) & swap & 11.39 [11.36, 11.42] & 0.115 [0.112, 0.118] & 0.152 [0.136, 0.167] \\
 & ref. & 22.69 [22.67, 22.72] & 0.917 [0.916, 0.918] & 0.793 [0.776, 0.811] \\
\midrule
all steps (0--38) & swap & 10.19 [10.14, 10.23] & 0.087 [0.085, 0.089] & 0.171 [0.155, 0.188] \\
\bottomrule
\end{tabular}
\end{table}

\section{Comparison between PLD Accountant and Edgeworth Approximation}\label{app:pld_vs_ew}

\subsection{Accounting Procedure}
\label{app:accounting_procedure}

\paragraph{Shared steps.}
(1)~Compute $\mu_t$ for every private step and $\bar{\mu}_{\mathcal{B}}=(\sum_{t\in\mathcal{B}}\mu_t^2)^{1/2}$ (\tabref{privacy_accounting} lists $\bar{\mu}_{\mathcal{B}}$ for each handoff).
(2)~Form the one-output pair $P_{\mathcal{B}}=\mathcal{N}(0,1)$, $Q_{\mathcal{B}}=(1-q_{\mathcal{B}})\mathcal{N}(0,1)+q_{\mathcal{B}}\mathcal{N}(\bar{\mu}_{\mathcal{B}},1)$ and its privacy loss $L_{\mathcal{B}}$ (Proposition~\ref{prop:output_subsampling}), in both directions ($Q_{\mathcal{B}}$ against $P_{\mathcal{B}}$ and $P_{\mathcal{B}}$ against $Q_{\mathcal{B}}$).
For a pair $(A,A')$ with privacy loss $\ell=\log(\mathrm{d}A/\mathrm{d}A')$, the hockey-stick divergence is $H_{e^\varepsilon}(A\|A')=\mathbb{E}_{A}[(1-e^{\varepsilon-\ell})_+]$, so $\delta(\varepsilon)$ depends on the pair only through the distribution of $\ell$ under $A$, its PLD.

\paragraph{PLD accountant (certified, reported as $\varepsilon_u$).}
(3)~For each direction, discretize the PLD of one output pessimistically on a grid of width $h=10^{-5}$, so that the hockey-stick divergence of the discrete PLD upper-bounds the true one at every $\varepsilon$. Tail mass below $e^{-50}$ is truncated and added to $\delta$.
(4)~Compose the $N_{\mathrm{out}}$ independent outputs by convolving the discretized PLD with itself $N_{\mathrm{out}}$ times, computed with the FFT. Any mass truncated during composition is also added to $\delta$. The result is a pessimistic PLD of $S_{\mathrm{out}}$.
(5)~For each direction evaluate $\delta(\varepsilon)=\sum_{s}\omega(s)(1-e^{\varepsilon-s})_+$ plus the truncated mass, where $\omega$ is the composed discrete PLD, take the larger of the two directions, and find the smallest $\varepsilon$ with $\delta_{\mathrm{out}}(\varepsilon)\le\delta$. This is $\varepsilon_u$.
(6)~Repeating (3)--(5) with every value rounded down gives $\varepsilon_\ell$, which is not a bound and only shows how much of $\varepsilon_u$ is discretization slack.
We use the implementation of \citet{doroshenko2022connect} (Google \texttt{dp\_accounting}) to automatically calculate these values given $\bar{\mu}_{\mathcal{B}}$ and subsampling rate.

\paragraph{Edgeworth approximation (estimate).}
(3$'$)~For $D\in\{P,Q\}$, compute the first four cumulants $\kappa^D_{r,\mathcal{B}}$ of $L_{\mathcal{B}}$ under $D_{\mathcal{B}}$ by numerical integration over $z$.
(4$'$)~Cumulants add over independent outputs, so $S_{\mathrm{out}}$ has $M_D=N_{\mathrm{out}}\kappa^D_{1,\mathcal{B}}$, $V_D=N_{\mathrm{out}}\kappa^D_{2,\mathcal{B}}$, $K_{3,D}=N_{\mathrm{out}}\kappa^D_{3,\mathcal{B}}$, and $K_{4,D}=N_{\mathrm{out}}\kappa^D_{4,\mathcal{B}}$ (Appendix~\ref{app:edgeworth_diagnostics}).
(5$'$)~Approximate $\Pr_D(S_{\mathrm{out}}>\varepsilon)\approx1-F_{\mathrm{EW}}\big((\varepsilon-M_D)/\sqrt{V_D}\big)$ with \eqref{edgeworth}, insert both probabilities into $\delta_{\mathrm{out}}(\varepsilon)$ of Proposition~\ref{prop:output_subsampling} for each direction, take the larger, and solve $\delta_{\mathrm{out}}(\varepsilon)=\delta$ by bisection.
(6$'$)~Check the diagnostics of Appendix~\ref{app:edgeworth_diagnostics}. The value is an estimate: the truncated series has an error of unknown sign and there is no discretization or truncation control.
The CLT column of \tabref{edgeworth_vs_clt} instead replaces the composed trade-off curve by a single Gaussian curve $G_\mu$, with $\mu$ from the summed first two moments of the per-output privacy loss~\citep{GDP}.

\subsection{Comparison}

The degree-two Edgeworth approximation used in the paper is a \textbf{numerical approximation}, not a one-sided certified DP upper bound. Although Propositions~\ref{prop:output_subsampling} and~\ref{prop:step_subsampling} exactly characterize the relevant block/per-step privacy-loss distributions, replacing their CDFs by a truncated Edgeworth expansion introduces an approximation error whose sign is not controlled in our finite heterogeneous setting.

To provide a formal guarantee, we additionally composed the mechanism actually implemented---within a private block, conditional per-step Gaussian privacy accumulates to $\bar{\mu}_{\mathcal B}=(\sum_{t\in\mathcal B}\mu_t^2)^{1/2}$, subsampling amplification is applied \textbf{once} to the block, and the result is composed over $N_{\mathrm{out}}$ released samples---using PLD~\citep{doroshenko2022connect} and RDP~\citep{wang2019subsampled} accountants. All accountants use the sampling model of Section~\ref{sec:subsampling} ($q=B/n$, add/remove adjacency), block parameter $\bar{\mu}_{\mathcal B}$, and the same output count. RDP is a conservative certified relaxation; the PLD method provides a rigorous upper bound and a lower bound that bracket the optimal privacy value after controlling discretization and truncation error. At $\delta=10^{-5}$, $N_{\mathrm{out}}=50{,}000$, $B=25$ for CIFAR-10, PLD grid spacing $h=10^{-5}$ and \texttt{log\_mass\_truncation\_bound} $=-50$:

\begin{table}[t]
\centering
\caption{Privacy accounting comparison for different empirical block sizes.}
\label{tab:privacy_accounting}
\scalebox{0.80}{
\begin{tabular}{rccccc}
\toprule
Empirical block (\%) & $\bar{\mu}_{\mathcal{B}}$ & Edgeworth estimate $\varepsilon_{\mathrm{ew}}$ & PLD lower bound & PLD upper bound & RDP upper bound \\
\midrule
55 & 0.69 & 0.2987 & 0.04880 & 0.2988 & 0.36 \\
62 & 1.56 & 1.68   & 1.44    & 1.69   & 2.63 \\
65 & 2.09 & 5.33   & 6.05    & 6.29   & 7.81 \\
67 & 2.84 & 23.0   & 25.49   & 25.7   & 48.8 \\
\bottomrule
\end{tabular}}
\end{table}

The comparison reveals two regimes in these evaluated configurations. For $\varepsilon \lesssim 2$, the Edgeworth estimate lies within $1\%$ of the certified PLD upper bound (55\% and 62\% blocks). For $\varepsilon \gtrsim 5$, it is \textbf{anti-conservative}: at the 65\% and 67\% blocks, the Edgeworth estimate falls below even the optimistic PLD \emph{lower} estimate (5.33 vs.\ $[6.05,\,6.29]$ and 22.98 vs.\ $[25.49,\,25.70]$), understating the certified value by about $10$--$15\%$.

\section{Differential Privacy and Gaussian Differential Privacy}

\subsection{Differential Privacy}\label{sec:appendix_dp}

The concept of differential privacy (DP)~\citep{DP_SGD, dwork2006calibrating} provides a formal guarantee that a model behaves almost identically regardless of whether any particular data sample is included in the training set. In other words, DP bounds the extent to which a single sample can influence the model's behavior. More precisely, a mechanism $\mathcal{M}$, which may be a training algorithm such as Stochastic Gradient Descent (SGD) or, more generally, any randomized algorithm including diffusion sampling, is said to be $\epsilon$-DP if the privacy loss is bounded by $\epsilon$: $\log{\frac{Pr[\mathcal{M}(\mathcal{D})=o]}{Pr[\mathcal{M}(\mathcal{D}')=o]}}\le\epsilon$, where $\mathcal{D}$ and $\mathcal{D}'$ differ in the record of a single individual; and $\mathcal{M}(\mathcal{D})=o$ denotes that applying the mechanism $\mathcal{M}$ on a dataset $\mathcal{D}$ ends up converging to the output event $o$.

For instance, if $\mathcal{M}$ is a training optimizer such as SGD, then $\mathcal{M}(\mathcal{D})$ is the trained model and $Pr(\mathcal{M}(\mathcal{D})=o)$ depicts the probability the model $o$ being the trained one. Intuitively, if deleting a single data point in $\mathcal{D}$ to obtain $\mathcal{D}'$ does not substantially change the probability of obtaining the same output model $o$, then the training algorithm can be regarded as privacy-preserving\footnote{SGD and Adam are not DP algorithms as the gradient update may be dominated by a huge outlier that significantly affects the model's parameters. As a result, a model trained with SGD/Adam can potentially reveal information about such outliers, leaving the data vulnerable to extraction attacks.}. Put more simply, if the probability the model converges to the same parameters is relatively equivalent regardless of whether a particular sample is present, then that sample cannot be said to meaningfully affect training. More generally, $\mathcal{M}$ is $(\varepsilon,\delta)$-DP if for all neighboring $\mathcal{D},\mathcal{D}'$ and every measurable set of outputs $O$,
\begin{align}\label{eq:approx_dp}
\Pr[\mathcal{M}(\mathcal{D})\in O] \le e^{\varepsilon}\Pr[\mathcal{M}(\mathcal{D}')\in O] + \delta.
\end{align}
The privacy-loss condition $\Pr\big[\log\frac{\Pr[\mathcal{M}(\mathcal{D})=o]}{\Pr[\mathcal{M}(\mathcal{D}')=o]}\ge\varepsilon\big]\le\delta$ implies \eqref{approx_dp} but is not equivalent to it.

A natural building block for designing private (either training or sampling) algorithms is the \emph{Gaussian mechanism}. This is particularly relevant for our setting, since each step in diffusion sampling injects Gaussian noise. Formally, given a function $\vh:\mathcal{D}\mapsto\mathbb{R}^p$, 
the \emph{Gaussian mechanism} releases
\begin{align}\label{eq:appendix_gaussian_mechanism}
   \mathcal{M}_{\mathrm{Gauss}}(\mathcal{D};\vh)
   = \vh(\mathcal{D}) + \bm{n},
\end{align}
where $\bm{n}\sim\mathcal{N}(0,\sigma^{2}\Delta_{\vh}^{2}\mathbf{I})$. Here, $\Delta_\vh$ denotes the \emph{global sensitivity}---the maximum $L_2$ difference between outputs on two neighboring datasets---and $\sigma$ controls the noise level as a function of $(\epsilon,\delta)$. By adding noise calibrated to sensitivity, the Gaussian mechanism ensures that the presence or absence of any single sample only has a limited influence on the output. This mechanism underlies differentially private training methods such as DP-SGD~\citep{DP_SGD}, and, as we will show, it also directly connects to the diffusion sampling process we analyze in this work.

As an interesting fact in differential privacy, it is widely known that subsampling amplifies privacy \citep{doi:10.1137/090756090}. 
For example, if we flip a coin to decide whether each entry of the training data is in or out to form a smaller dataset, we would expect the mechanism using the smaller dataset to be twice as private as the original mechanism using the full dataset. With a flipping probability of $50\%$, every individual benefits from perfect privacy if the individual is not included in the smaller dataset. To benefit from the privacy amplification effect, a subsampled Gaussian mechanism (applying the Gaussian mechanism on randomly subsampled data) is also widely used \citep{DP_SGD, wang2019subsampled}. We will also study the effect of subsampling on the privacy loss incurring during the diffusion sampling process.

\subsection{Gaussian Differential Privacy}\label{sec:appendix_gdp}

While the Gaussian mechanism provides a natural foundation for analyzing privacy, the classical $(\epsilon,\delta)$-DP framework yields only loose bounds when these mechanisms are composed, often resulting in substantially looser guarantees than the actual privacy loss. To address this, we adopt the more refined \emph{f}-DP framework~\citep{GDP}, which characterizes privacy through hypothesis testing and provides an exact trade-off curve (function), rather than the looser worst-case bounds of $(\epsilon,\delta)$-DP.

A particularly important special case is \emph{Gaussian Differential Privacy (GDP)}, which describes the privacy profile of the Gaussian mechanism exactly using a single parameter $\mu$. For example, in the case of \eqref{appendix_gaussian_mechanism} with $\bm{n}\sim\mathcal{N}(0,\sigma^{2}\Delta_{\vh}^{2}\mathbf{I})$, the Gaussian mechanism corresponds to $\frac{1}{\sigma}$-GDP. 
For a Gaussian mechanism, we can define a trade-off function $T$ using the standard normal CDF $\Phi$ given by, 
\begin{align}\label{eq:appendix_trade_off_Gaussian}
    G_\mu:= T(\Nrm(0,1), \Nrm(\mu,1)).  \quad G_\mu(\alpha):= \Phi(\Phi^{-1}(1-\alpha)-\mu).
\end{align}

This makes GDP especially well aligned with diffusion sampling, where Gaussian perturbations are intrinsic to every denoising step. Moreover, GDP is closed under composition (Corollary 2 of \citet{GDP}): the cumulative privacy loss of $n$ mechanisms with parameters $\mu_1,\ldots,\mu_n$ is given tightly by
\begin{align}\label{eq:appendix_gdp_composition}
    \mu=\sqrt{\sum_{i=1}^{n}\mu_{i}^{2}}.
\end{align}
Thus, GDP not only matches the Gaussian mechanism exactly but also enables precise accounting of privacy loss across many Gaussian steps that constitute diffusion sampling. A $\mu$-GDP mechanism is $(\varepsilon,\delta(\varepsilon))$-DP for every $\varepsilon\ge0$ with the closed form~\citep{GDP}
\begin{align}\label{eq:gdp_to_dp}
    \delta(\varepsilon) = \Phi\left(-\frac{\varepsilon}{\mu} + \frac{\mu}{2}\right) - e^{\varepsilon} \Phi\left(-\frac{\varepsilon}{\mu} - \frac{\mu}{2}\right),
\end{align}
which we use for all full-batch conversions.

For subsampled Gaussian mechanisms,  the trade-off function in \eqref{appendix_trade_off_Gaussian} needs to be modified to incorporate the randomness coming from the subsampling process and the GDP composition in \eqref{appendix_gdp_composition} is no longer valid. In this case, one can benefit from using the central limit theorem, which tells us that composing many $f$-DP mechanisms converges to GDP, as formally stated in Theorem 3.5 of \citet{GDP}.

\section{Full-batch sampling}\label{app:fullbatch}

\begin{figure}[t]
\centering
    \includegraphics[width=0.5\linewidth]{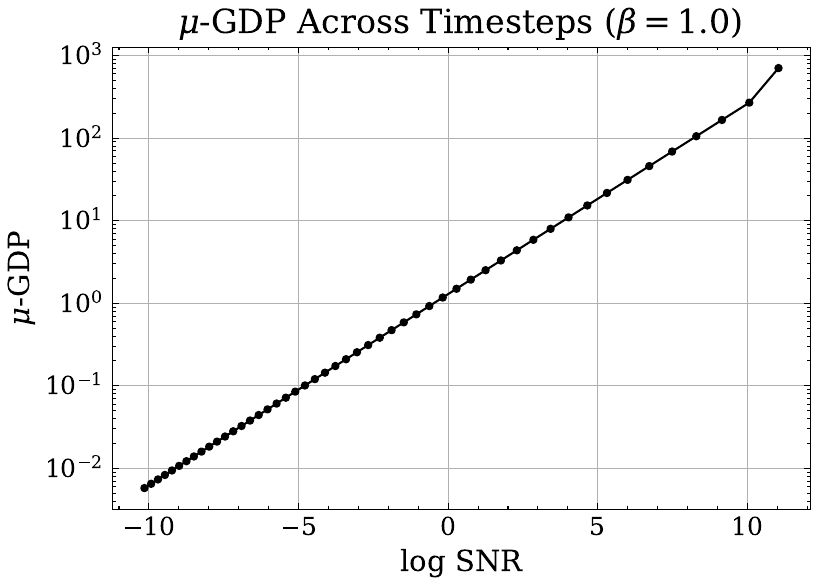}
    \caption{$\mu$-GDP for each time step with a full batch of data.}
    \label{fig:placeholder}
\end{figure}

In the full-batch regime every record contributes to every denoising step, so the per-step parameters $\mu_t$ compose exactly by \eqref{gdp_composition} and convert to $(\varepsilon,\delta)$ by \eqref{gdp_to_dp}. We vary the minimum noise level $\sigma_{\min}$ to understand its impact on privacy guarantees. Figure~\ref{fig:placeholder} shows how the per-step $\mu$ grows as the noise level falls.

\begin{figure*}[t]
    \centering
    \begin{subfigure}[t]{0.30\textwidth}
        \centering
        \includegraphics[width=1\textwidth]{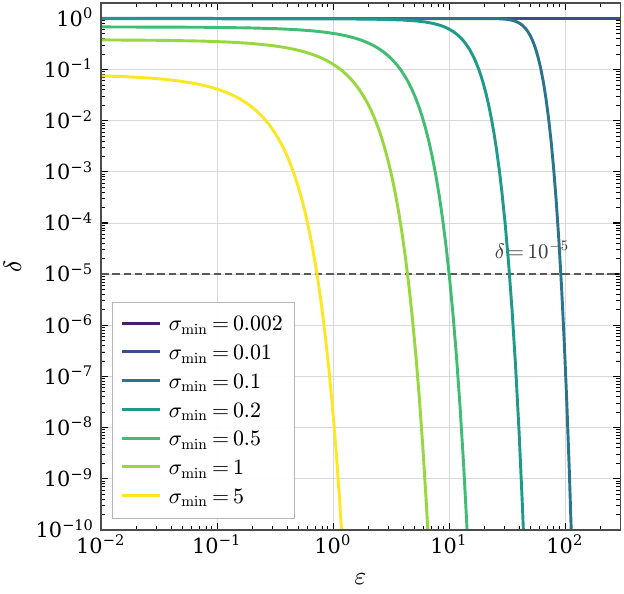}
    \end{subfigure}
    \hfill
    \begin{subfigure}{0.33\textwidth}
        \centering
        \includegraphics[width=0.9\textwidth]{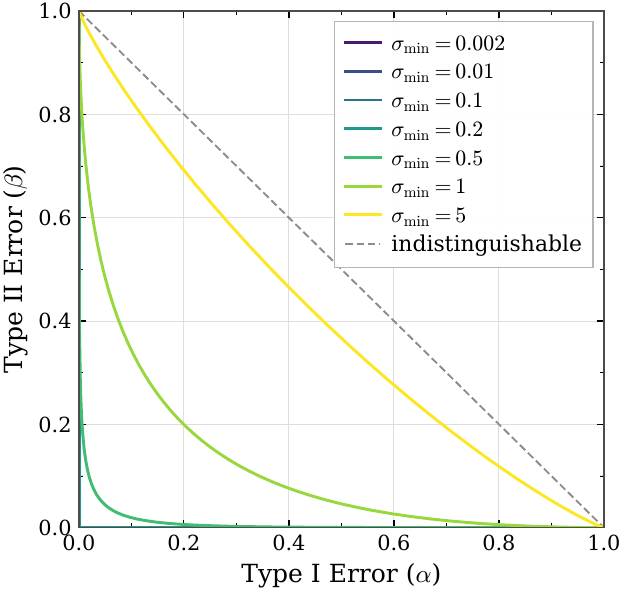}
    \end{subfigure}
    \hfill
    \begin{subfigure}[t]{0.35\textwidth}
        \centering
        \includegraphics[width=\textwidth]{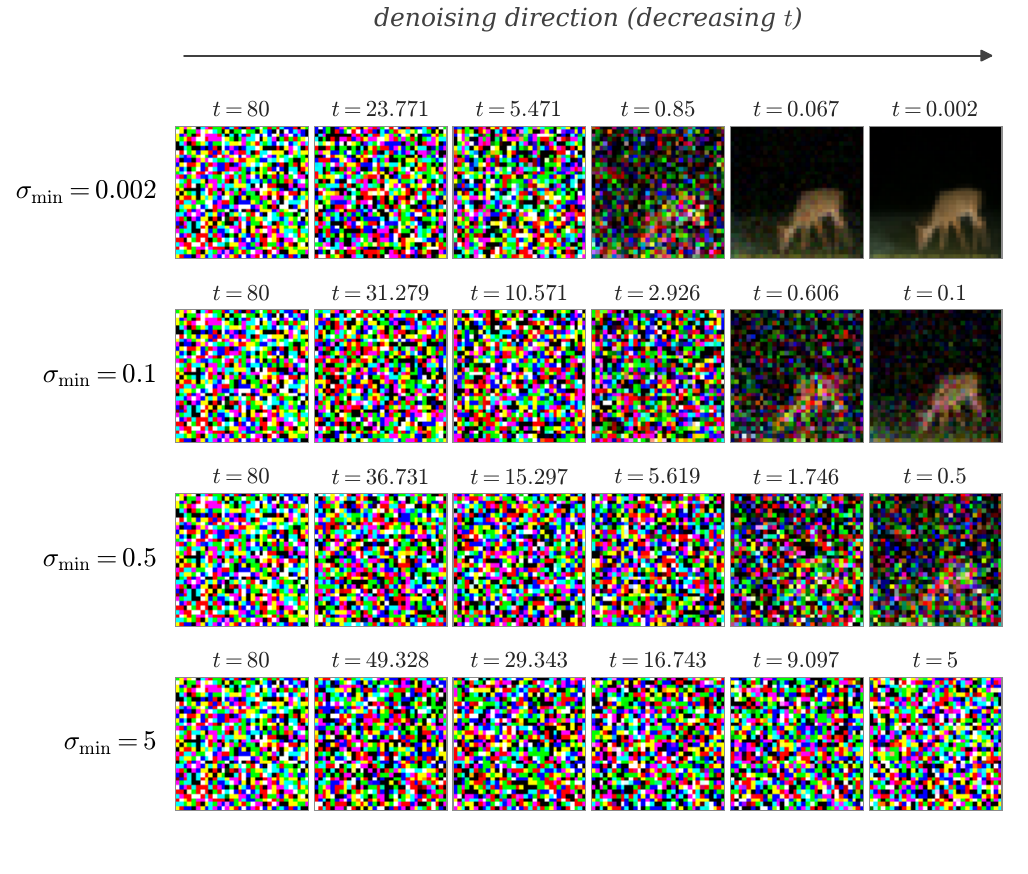}
    \end{subfigure}
    \caption{GDP Analysis of Diffusion-Based Sampling (full batch) and Corresponding Diffusion Process Visualization. The noise scheduling constants are $\sigma_{max} = 80$, $N=50$, $\beta = 1$,  and $\rho = 7$. The $(\epsilon, \delta)$-DP guarantees account for a single sample. \textbf{Left.} GDP conversion to $(\epsilon, \delta)$-DP for varying $\sigma_{\min}$. \textbf{Middle.} Type I and Type II Error under full-batch GDP with varying $\sigma_{min}$. \textbf{Right.} Visualization of Diffusion Process with Varying $\sigma_{\min}$. Only six timesteps are shown for figure clarity. }
    \label{fig:gdp_sigma_min}
\end{figure*}

\paragraph{Lower $\sigma_{\min}$ corresponds to less privacy.} Figure~\ref{fig:gdp_sigma_min} shows how the choice of $\sigma_{\min}$ affects the privacy-utility trade-off. As $\sigma_{\min}$ decreases (getting closer to the training data), the GDP analysis gives worse $(\varepsilon,\delta)$-DP guarantees.

\paragraph{The full-batch empirical denoiser does not give good utility for its privacy cost.} At $\varepsilon \approx 1$ (corresponding to $\sigma_{\min}=5$), the generated sample is hard to distinguish from pure noise (Figure~\ref{fig:gdp_sigma_min}, right). For a recognizable sample we need $\sigma_{\min}=0.1$, which corresponds to $\varepsilon \approx 200$.

\paragraph{Scalability of full-batch sampling.} Even at low $\sigma_{\min}$, full-batch diffusion sampling does not scale well with the number of generated samples. Since each denoising step operates on the full dataset, the per-sample privacy loss is already high, and composing multiple samples results in a rapidly growing total privacy budget under GDP. Subsampling alleviates this through privacy amplification (Section~\ref{sec:subsampling}).

\subsection{Privacy loss in relation to critical windows}
\begin{figure*}[t]
    \centering
    \begin{subfigure}{0.32\textwidth}
        \centering
        \includegraphics[width=0.9\textwidth]{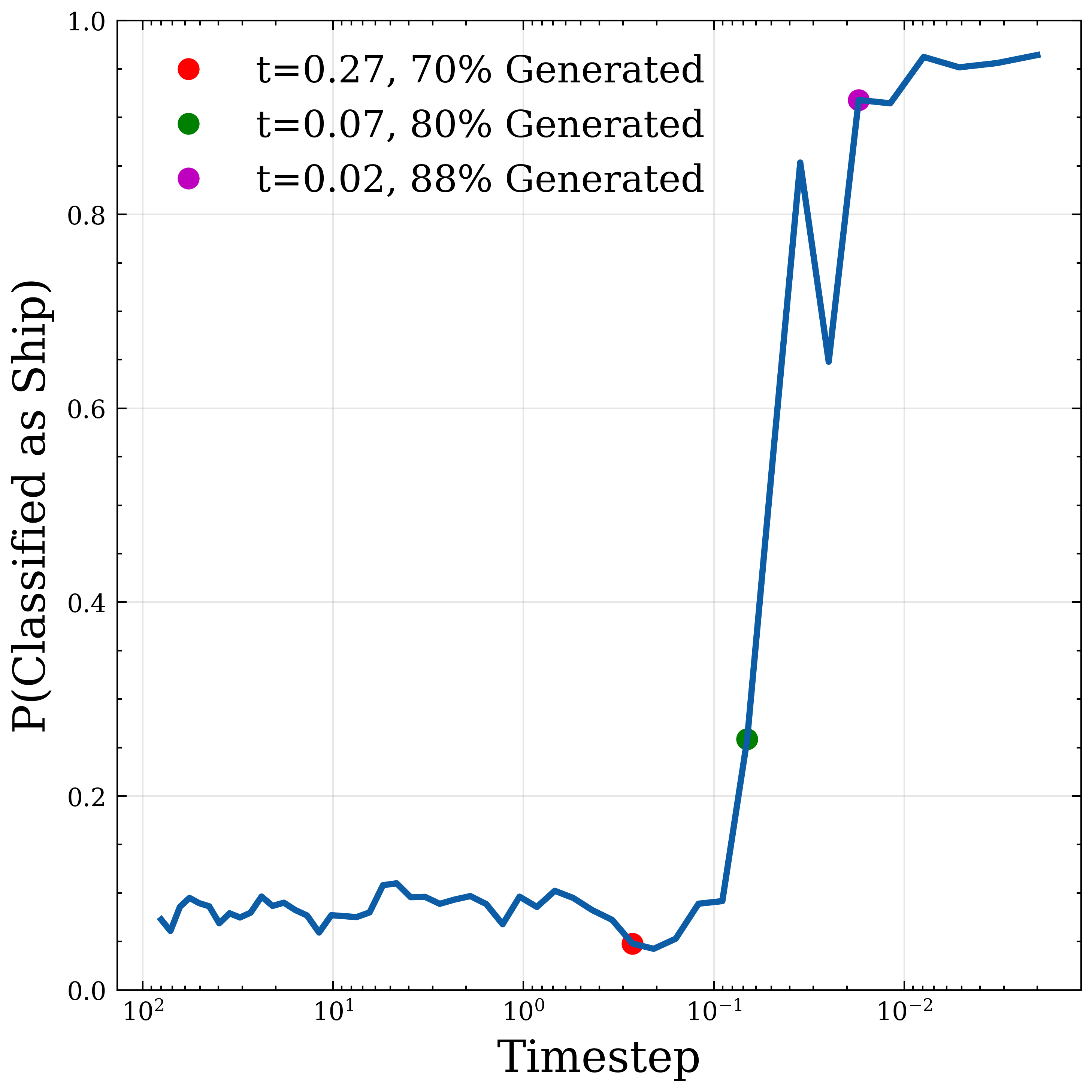}
        \caption{}
    \end{subfigure}
    \hfill
    \begin{subfigure}{0.32\textwidth}
        \centering
        \includegraphics[width=0.9\textwidth]{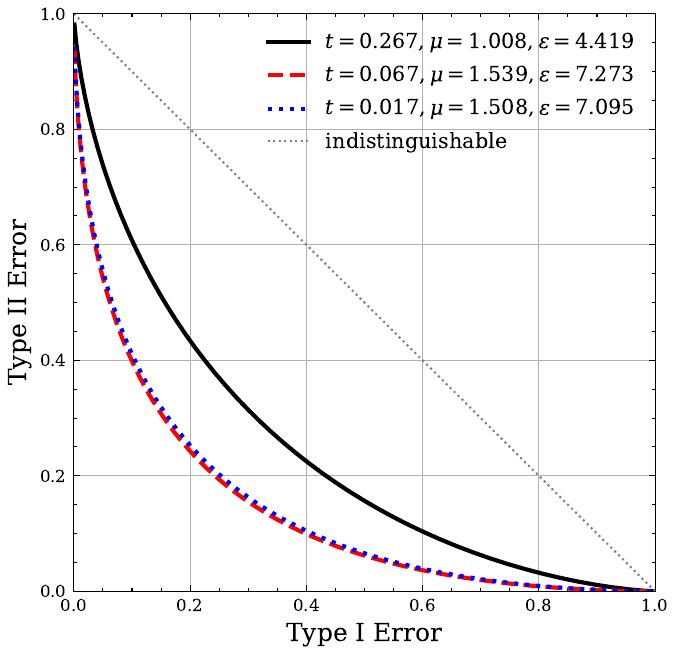}
        \caption{}
    \end{subfigure}
    \hfill
    \begin{subfigure}{0.33\textwidth}
        \centering
        \includegraphics[width=0.9\textwidth]{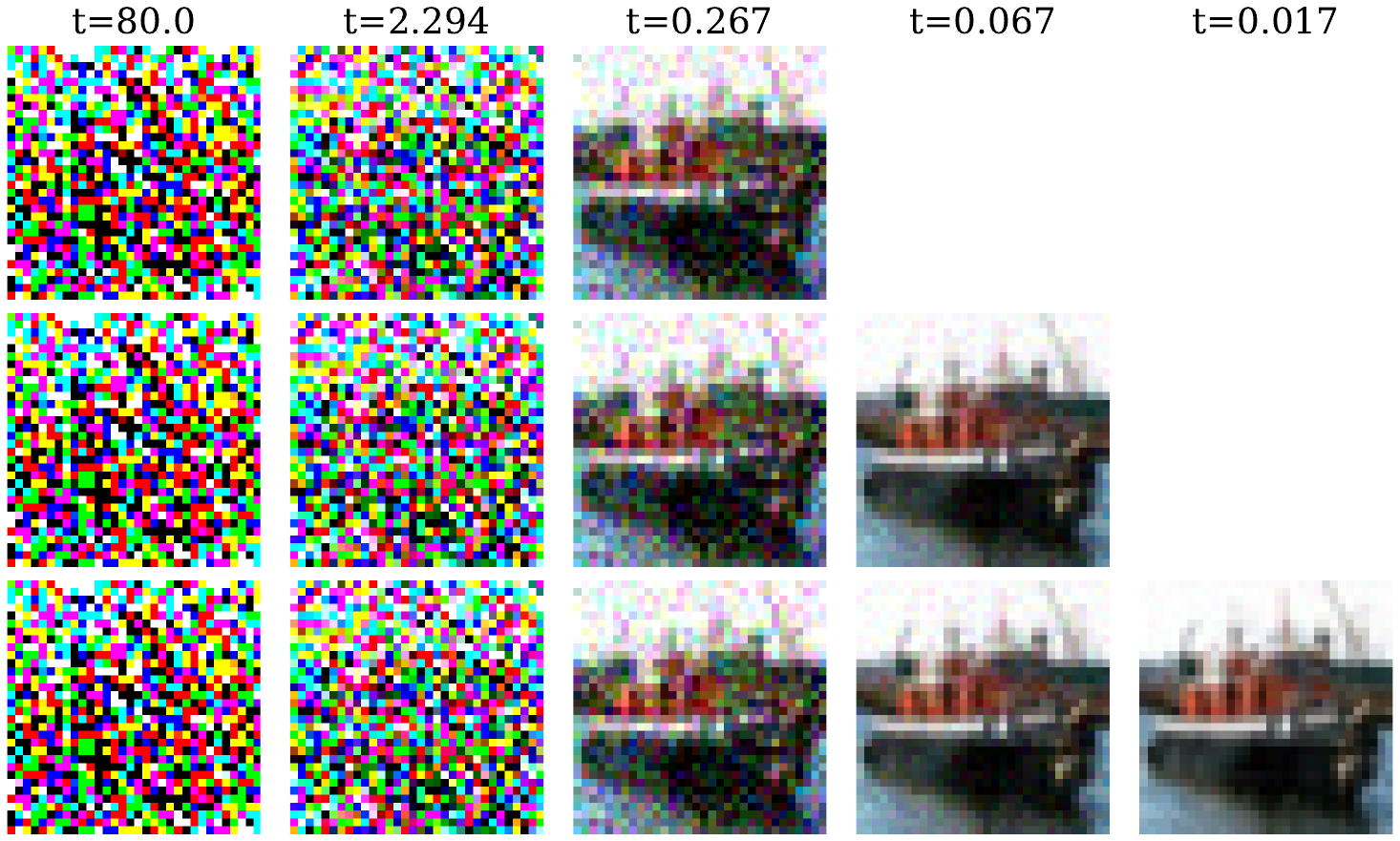}
        \caption{}
    \end{subfigure}
    \caption{Detection of critical window. \textbf{(a)} zero-shot CLIP model detects the probability of an image being generated belonging to the \textit{ship} class. \textbf{(b)} Privacy costs at three time points. \textbf{(c)} Generated images denoised at varying timesteps.}
    \label{fig:main}
\end{figure*}

\begin{figure}[t]
    \centering
    \begin{subfigure}{0.47\textwidth}
        \centering
        \includegraphics[width=\textwidth]{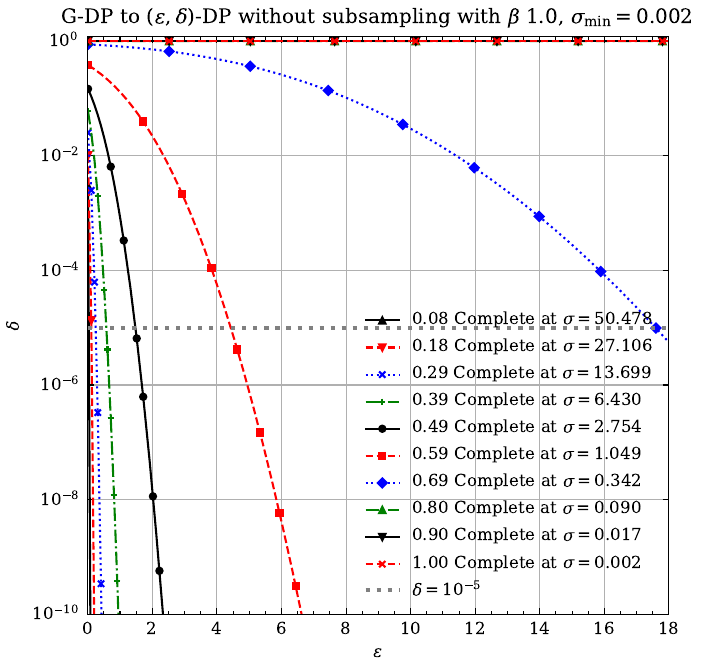}
        \label{fig:example}
    \end{subfigure}
    \hfill
    \begin{subfigure}{0.47\textwidth}
        \centering
        \includegraphics[width=\textwidth]{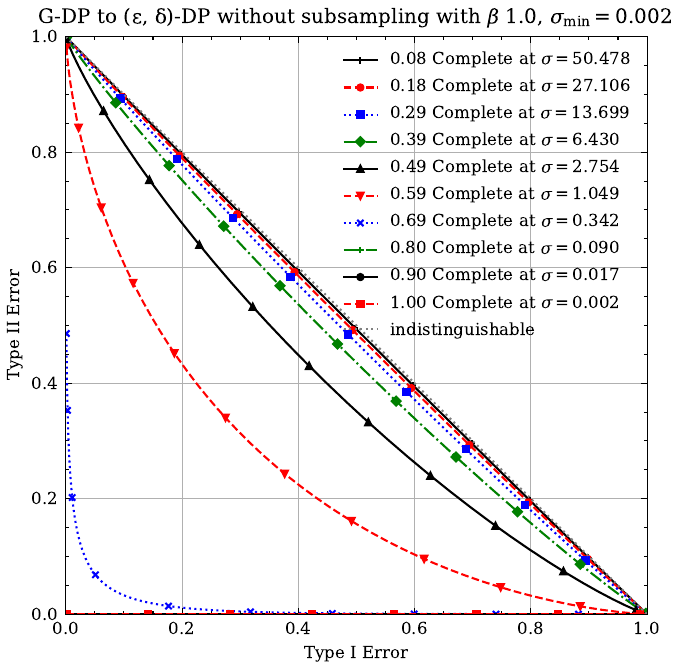}
        \label{fig:sub2}
    \end{subfigure}
    \caption{$(\epsilon, \delta)$-DP traces (Left) and Type I and II error profiles (Right) under GDP with varying stopping time points in the reverse process}
    \label{fig:critical_window_full}
\end{figure}

Figure~\ref{fig:main} shows the CLIP-based detection of the critical window used in Section~\ref{subsec:privacy_critical_windows}, and Figure~\ref{fig:critical_window_full} shows $(\epsilon, \delta)$-DP traces (left) and Type I and II error profiles (right) under GDP with varying stopping time points in the reverse process.

\section{Evaluation of our Hybrid Empirical Denoiser and Neural Denoiser on CIFAR-10 and CelebA} \label{app:hybrid_evaluation}

The CIFAR-10 experiment uses an ImageNet-pretrained diffusion model for the neural denoiser, and we adopt the $\rho$-power schedule of EDM~\citep{karras2022elucidating}. The scheduler is parameterized by the minimum and maximum noise levels $(\sigma_{min}, \sigma_{max})$, the noise decay rate $\rho$, and the number of denoising steps $N$. We experiment with $\sigma_{\min}=0.002$, $\sigma_{\max}=80$, $\rho=7$, $N=39$, and the public normalization $\beta=1$ (Appendix~\ref{app:beta-ablation}).

We conduct a hyperparameter search over the number of re-noising steps $T \in \{1, 3, 5, 7, 9\}$,
empirical generation proportion $prop \in \{55, 62, 65, 67, 70\}\%$, and subsample batch size $B \in \{25, 50, 100, 200\}$, giving 95 configurations (the 70\%, $B=200$ setting is not reported due to overflowing privacy calculations), with results in Table~\ref{tab:hyperparam_search_rowsubsample}. For SDEdit with output-wise subsampling of size $B$, we run empirical denoising for the first $prop\%$ of the generation process, up to $t_0$. Afterwards, we re-noise by adding Gaussian noise with scale $(\sigma_{t_1}^2 - \sigma_{t_0}^2)^{1/2}$. We continue denoising with the public Imagenet denoiser starting at $t_1$ until the end. We generate 50,000 samples for every run and compute the FID against CIFAR10.

Experiments were conducted on NVIDIA L40S GPUs with 48GB memory. Since our method primarily performs private sampling rather than full diffusion training, compute requirements were moderate. CIFAR-10 experiments typically required approximately 2.5 GPU-hours per run.

For CelebA~\citep{liu2015faceattributes} at $64 \times 64$ ($n = 162{,}770$, $\delta = 10^{-6}$), we use FFHQ~\citep{karras2019style}, trained without access to the private CelebA training split, as the disjoint public source for the neural denoiser, and keep the same EDM scheduler ($\sigma_{min} = 0.002$, $\sigma_{max} = 80$, $\rho = 7$, $N = 39$) and public normalization $\beta = 1$. We sweep the same base grid of re-noising steps $T \in \{1, 3, 5, 7, 9\}$, empirical generation proportion $prop \in \{55, 62, 65, 67, 70\}\%$, and subsample batch size $B \in \{25, 50, 100, 200\}$, and at $T = 3$ additionally refine $B$ (to $\{60, 70\}$ for $prop \in \{62, 65\}\%$ and to $\{80, 90, 120, 140\}$ for $prop \in \{65, 67\}\%$), giving 112 configurations in total, with results reported in Table~\ref{tab:celeba_hyperparam_main}. The pipeline is otherwise identical to the CIFAR-10 setting: we run empirical denoising for the first $prop\%$ of generation up to $t_0$, re-noise by adding Gaussian noise with scale $(\sigma_{t_1}^2 - \sigma_{t_0}^2)^{1/2}$, and complete generation with the public FFHQ denoiser from $t_1$ until the end. We generate 50,000 samples for every run and compute the FID against CelebA. The Edgeworth diagnostic $\varepsilon_{sym}$ is available only for the refined $T = 3$ rows and is marked invalid where the reuse-validity flag is false or a cumulant ratio $\rho \geq 1$ (e.g., $prop = 65\%$, $B = 60$, where $\rho^Q_4 = 1.19$). Table~\ref{tab:celeba_seeds} repeats the $B = 100$, $T = 3$ handoff sweep over three generation seeds.

\renewcommand{\arraystretch}{0.85}
\setlength{\tabcolsep}{6pt}

\begin{footnotesize}
\begin{longtable}{cccrcl}
  \caption{%
    Hyperparameter search for output-wise subsample SDEdit on CIFAR-10.
    We sweep the number of re-noising steps $T \in \{1, 3, 5, 7, 9\}$,
    the proportion of empirical (denoiser) generation
    $p \in \{55, 62, 65, 67, 70\}\%$,
    and the subsample batch size $B \in \{25, 50, 100, 200\}$; $95$ configurations in total.
    FID measures sample quality (lower is better).
    $\mathrm{PLD}\,[\varepsilon_{\ell}, \varepsilon_{u}]$ is the PLD interval under the
    sampling model of Section~\ref{sec:subsampling} (optimistic lower / pessimistic upper
    estimate, Google \texttt{dp\_accounting}). Only $\varepsilon_u$ is a certified bound.
    $\varepsilon_{ew}$ is the second degree Edgeworth approximation
    (lower is better).
  }
  \label{tab:hyperparam_search_rowsubsample} \\
  \toprule
  \textbf{Noise Steps} & \textbf{Prop.\ Empirical (\%)} & \textbf{Subsample Batch Size} & \textbf{FID} $\downarrow$ & \textbf{PLD} $[\bm{\varepsilon}_{\ell}, \bm{\varepsilon}_{u}]$ $\downarrow$ & $\bm{\varepsilon}_{ew}$ $\downarrow$ \\
  \midrule
  \endfirsthead
  \multicolumn{6}{l}{\small\textit{Table~\ref{tab:hyperparam_search_rowsubsample} continued}} \\[2pt]
  \toprule
  \textbf{Noise Steps} & \textbf{Prop.\ Empirical (\%)} & \textbf{Subsample Batch Size} & \textbf{FID} $\downarrow$ & \textbf{PLD} $[\bm{\varepsilon}_{\ell}, \bm{\varepsilon}_{u}]$ $\downarrow$ & $\bm{\varepsilon}_{ew}$ $\downarrow$ \\
  \midrule
  \endhead
  \midrule
  \multicolumn{6}{r}{\small\textit{Continued on next page}} \\
  \endfoot
  \bottomrule
  \endlastfoot
  1 &  55 &  25 &  147.43 & $[0.049,\,0.299]$      &    0.299 \\
    &  55 &  50 &  143.57 & $[0.386,\,0.636]$      &    0.636 \\
    &  55 & 100 &  141.23 & $[1.111,\,1.361]$      &    1.360 \\
    &  55 & 200 &  139.74 & $[2.702,\,2.952]$      &    2.946 \\
    &  62 &  25 &   78.79 & $[1.442,\,1.691]$      &    1.683 \\
    &  62 &  50 &   78.23 & $[3.172,\,3.422]$      &    3.513 \\
    &  62 & 100 &   78.73 & $[7.009,\,7.259]$      &    7.515 \\
    &  62 & 200 &   80.37 & $[16.009,\,16.259]$    &   16.706 \\
    &  65 &  25 &   78.53 & $[6.052,\,6.289]$      &    5.331 \\
    &  65 &  50 &   77.65 & $[11.136,\,11.381]$    &   10.777 \\
    &  65 & 100 &   78.24 & $[22.530,\,22.780]$    &   22.540 \\
    &  65 & 200 &   79.77 & $[49.332,\,49.582]$    &   49.625 \\
    &  67 &  25 &   80.33 & $[25.486,\,25.696]$    &   22.982 \\
    &  67 &  50 &   80.59 & $[46.196,\,46.438]$    &   44.481 \\
    &  67 & 100 &   81.94 & $[90.634,\,90.898]$    &   89.733 \\
    &  67 & 200 &   84.24 & $[190.673,\,190.915]$  &  190.198 \\
    &  70 &  25 &   74.98 & $[108.154,\,108.444]$  &  104.035 \\
    &  70 &  50 &   76.42 & $[194.318,\,194.645]$  &  191.700 \\
    &  70 & 100 &   79.52 & $[369.696,\,370.025]$  &  368.188 \\
  \midrule
  3 &  55 &  25 &   45.67 & $[0.049,\,0.299]$      &    0.299 \\
    &  55 &  50 &   46.09 & $[0.386,\,0.636]$      &    0.636 \\
    &  55 & 100 &   47.69 & $[1.111,\,1.361]$      &    1.360 \\
    &  55 & 200 &   49.92 & $[2.702,\,2.952]$      &    2.946 \\
    &  62 &  25 &   29.66 & $[1.442,\,1.691]$      &    1.683 \\
    &  62 &  50 &   31.51 & $[3.172,\,3.422]$      &    3.513 \\
    &  62 & 100 &   34.59 & $[7.009,\,7.259]$      &    7.515 \\
    &  62 & 200 &   38.46 & $[16.009,\,16.259]$    &   16.706 \\
    &  65 &  25 &   25.00 & $[6.052,\,6.289]$      &    5.331 \\
    &  65 &  50 &   27.13 & $[11.136,\,11.381]$    &   10.777 \\
    &  65 & 100 &   31.01 & $[22.530,\,22.780]$    &   22.540 \\
    &  65 & 200 &   35.12 & $[49.332,\,49.582]$    &   49.625 \\
    &  67 &  25 &   20.49 & $[25.486,\,25.696]$    &   22.982 \\
    &  67 &  50 &   23.27 & $[46.196,\,46.438]$    &   44.481 \\
    &  67 & 100 &   27.48 & $[90.634,\,90.898]$    &   89.733 \\
    &  67 & 200 &   32.59 & $[190.673,\,190.915]$  &  190.198 \\
    &  70 &  25 &   16.38 & $[108.154,\,108.444]$  &  104.035 \\
    &  70 &  50 &   19.46 & $[194.318,\,194.645]$  &  191.700 \\
    &  70 & 100 &   24.28 & $[369.696,\,370.025]$  &  368.188 \\
  \midrule
  5 &  55 &  25 &   33.16 & $[0.049,\,0.299]$      &    0.299 \\
    &  55 &  50 &   34.12 & $[0.386,\,0.636]$      &    0.636 \\
    &  55 & 100 &   36.19 & $[1.111,\,1.361]$      &    1.360 \\
    &  55 & 200 &   38.89 & $[2.702,\,2.952]$      &    2.946 \\
    &  62 &  25 &   29.39 & $[1.442,\,1.691]$      &    1.683 \\
    &  62 &  50 &   30.94 & $[3.172,\,3.422]$      &    3.513 \\
    &  62 & 100 &   33.94 & $[7.009,\,7.259]$      &    7.515 \\
    &  62 & 200 &   37.44 & $[16.009,\,16.259]$    &   16.706 \\
    &  65 &  25 &   27.35 & $[6.052,\,6.289]$      &    5.331 \\
    &  65 &  50 &   29.12 & $[11.136,\,11.381]$    &   10.777 \\
    &  65 & 100 &   32.55 & $[22.530,\,22.780]$    &   22.540 \\
    &  65 & 200 &   36.30 & $[49.332,\,49.582]$    &   49.625 \\
    &  67 &  25 &   24.68 & $[25.486,\,25.696]$    &   22.982 \\
    &  67 &  50 &   26.94 & $[46.196,\,46.438]$    &   44.481 \\
    &  67 & 100 &   30.70 & $[90.634,\,90.898]$    &   89.733 \\
    &  67 & 200 &   34.91 & $[190.673,\,190.915]$  &  190.198 \\
    &  70 &  25 &   22.65 & $[108.154,\,108.444]$  &  104.035 \\
    &  70 &  50 &   25.21 & $[194.318,\,194.645]$  &  191.700 \\
    &  70 & 100 &   29.07 & $[369.696,\,370.025]$  &  368.188 \\
  \midrule
  7 &  55 &  25 &   30.99 & $[0.049,\,0.299]$      &    0.299 \\
    &  55 &  50 &   31.97 & $[0.386,\,0.636]$      &    0.636 \\
    &  55 & 100 &   34.01 & $[1.111,\,1.361]$      &    1.360 \\
    &  55 & 200 &   36.38 & $[2.702,\,2.952]$      &    2.946 \\
    &  62 &  25 &   30.54 & $[1.442,\,1.691]$      &    1.683 \\
    &  62 &  50 &   31.81 & $[3.172,\,3.422]$      &    3.513 \\
    &  62 & 100 &   34.34 & $[7.009,\,7.259]$      &    7.515 \\
    &  62 & 200 &   37.34 & $[16.009,\,16.259]$    &   16.706 \\
    &  65 &  25 &   29.70 & $[6.052,\,6.289]$      &    5.331 \\
    &  65 &  50 &   31.26 & $[11.136,\,11.381]$    &   10.777 \\
    &  65 & 100 &   34.02 & $[22.530,\,22.780]$    &   22.540 \\
    &  65 & 200 &   37.18 & $[49.332,\,49.582]$    &   49.625 \\
    &  67 &  25 &   28.23 & $[25.486,\,25.696]$    &   22.982 \\
    &  67 &  50 &   29.86 & $[46.196,\,46.438]$    &   44.481 \\
    &  67 & 100 &   33.13 & $[90.634,\,90.898]$    &   89.733 \\
    &  67 & 200 &   36.41 & $[190.673,\,190.915]$  &  190.198 \\
    &  70 &  25 &   26.32 & $[108.154,\,108.444]$  &  104.035 \\
    &  70 &  50 &   28.51 & $[194.318,\,194.645]$  &  191.700 \\
    &  70 & 100 &   31.76 & $[369.696,\,370.025]$  &  368.188 \\
  \midrule
  9 &  55 &  25 &   30.43 & $[0.049,\,0.299]$      &    0.299 \\
    &  55 &  50 &   31.41 & $[0.386,\,0.636]$      &    0.636 \\
    &  55 & 100 &   33.14 & $[1.111,\,1.361]$      &    1.360 \\
    &  55 & 200 &   35.07 & $[2.702,\,2.952]$      &    2.946 \\
    &  62 &  25 &   30.71 & $[1.442,\,1.691]$      &    1.683 \\
    &  62 &  50 &   31.79 & $[3.172,\,3.422]$      &    3.513 \\
    &  62 & 100 &   33.94 & $[7.009,\,7.259]$      &    7.515 \\
    &  62 & 200 &   36.49 & $[16.009,\,16.259]$    &   16.706 \\
    &  65 &  25 &   30.30 & $[6.052,\,6.289]$      &    5.331 \\
    &  65 &  50 &   31.80 & $[11.136,\,11.381]$    &   10.777 \\
    &  65 & 100 &   34.00 & $[22.530,\,22.780]$    &   22.540 \\
    &  65 & 200 &   36.74 & $[49.332,\,49.582]$    &   49.625 \\
    &  67 &  25 &   29.97 & $[25.486,\,25.696]$    &   22.982 \\
    &  67 &  50 &   31.31 & $[46.196,\,46.438]$    &   44.481 \\
    &  67 & 100 &   33.87 & $[90.634,\,90.898]$    &   89.733 \\
    &  67 & 200 &   36.69 & $[190.673,\,190.915]$  &  190.198 \\
    &  70 &  25 &   29.25 & $[108.154,\,108.444]$  &  104.035 \\
    &  70 &  50 &   31.11 & $[194.318,\,194.645]$  &  191.700 \\
    &  70 & 100 &   33.78 & $[369.696,\,370.025]$  &  368.188 \\
\end{longtable}
\end{footnotesize}
\renewcommand{\arraystretch}{1.0}

\begin{footnotesize}
\begin{longtable}{cccrcl}
  \caption{Hyperparameter search for output-wise subsample SDEdit on \textbf{CelebA}. The base grid sweeps re-noising steps $T \in \{1,3,5,7,9\}$, proportion of empirical (denoiser) generation $p \in \{55,62,65,67,70\}\%$, and subsample batch size $B \in \{25,50,100,200\}$; at $T=3$ we additionally refine $B$ (to $\{60,70\}$ for $p\in\{62,65\}\%$ and to $\{80,90,120,140\}$ for $p\in\{65,67\}\%$), giving $112$ configurations. FID measures sample quality (lower is better). $\mathrm{PLD}\,[\varepsilon_{\ell},\varepsilon_{u}]$ is the PLD interval under the sampling model of Section~\ref{sec:subsampling} (optimistic lower / pessimistic upper estimate, Google \texttt{dp\_accounting}). Only $\varepsilon_u$ is a certified bound. $\varepsilon_{ew}$ is the (diagnostic) Edgeworth symmetric privacy loss; it is marked \textit{invalid} where the reuse validity flag is false or a cumulant ratio $\rho \ge 1$ (diagnostics available only for the refined $T=3$ rows).}
  \label{tab:celeba_hyperparam_main} \\
  \toprule
  \textbf{Noise Steps} & \textbf{Prop.\ Empirical (\%)} & \textbf{Subsample Batch Size} & \textbf{FID} $\downarrow$ & \textbf{PLD} $[\bm{\varepsilon}_{\ell}, \bm{\varepsilon}_{u}]$ $\downarrow$ & $\bm{\varepsilon}_{ew}$ $\downarrow$ \\
  \midrule
  \endfirsthead
  \multicolumn{6}{l}{\small\textit{Table~\ref{tab:celeba_hyperparam_main} continued}} \\[2pt]
  \toprule
  \textbf{Noise Steps} & \textbf{Prop.\ Empirical (\%)} & \textbf{Subsample Batch Size} & \textbf{FID} $\downarrow$ & \textbf{PLD} $[\bm{\varepsilon}_{\ell}, \bm{\varepsilon}_{u}]$ $\downarrow$ & $\bm{\varepsilon}_{ew}$ $\downarrow$ \\
  \midrule
  \endhead
  \midrule
  \multicolumn{6}{r}{\small\textit{Continued on next page}} \\
  \endfoot
  \bottomrule
  \endlastfoot
  1 &  55 &  25 &  116.96 & $[0.000,\,0.099]$      & 0.099 \\
    &  55 &  50 &  113.07 & $[0.000,\,0.208]$      & 0.208 \\
    &  55 & 100 &  109.95 & $[0.185,\,0.435]$      & 0.435 \\
    &  55 & 200 &  107.90 & $[0.666,\,0.916]$      & 0.915 \\
    &  62 &  25 &   45.45 & $[0.516,\,0.767]$      & 0.563 \\
    &  62 &  50 &   45.23 & $[1.179,\,1.430]$      & 1.150 \\
    &  62 & 100 &   46.24 & $[2.308,\,2.558]$      & 2.361 \\
    &  62 & 200 &   47.05 & $[4.654,\,4.904]$      & 4.912 \\
    &  65 &  25 &   37.32 & $[3.623,\,3.891]$      & \textit{invalid} ($\rho_4^{Q}{=}3.32$) \\
    &  65 &  50 &   38.20 & $[5.249,\,5.500]$      & \textit{invalid} ($\rho_4^{Q}{=}1.48$) \\
    &  65 & 100 &   39.87 & $[8.482,\,8.737]$      & 7.147 \\
    &  65 & 200 &   41.71 & $[15.522,\,15.769]$    & 14.552 \\
    &  67 &  25 &   30.23 & $[13.360,\,13.657]$    & \textit{invalid} ($\rho_4^{Q}{=}3.46$) \\
    &  67 &  50 &   32.07 & $[20.517,\,20.746]$    & \textit{invalid} ($\rho_4^{Q}{=}1.28$) \\
    &  67 & 100 &   34.65 & $[34.120,\,34.392]$    & 30.085 \\
    &  67 & 200 &   37.48 & $[61.908,\,62.120]$    & 58.631 \\
    &  70 &  25 &   24.58 & $[52.702,\,53.019]$    & \textit{invalid} ($\rho_4^{P}{=}2.26$) \\
    &  70 &  50 &   27.01 & $[83.289,\,83.470]$    & \textit{invalid} ($\rho_4^{P}{=}1.15$) \\
    &  70 & 100 &   30.48 & $[140.112,\,140.393]$  & 132.911 \\
    &  70 & 200 &   33.82 & $[251.533,\,251.653]$  & 245.964 \\
  \midrule
  3 &  55 &  25 &   35.16 & $[0.000,\,0.099]$      & 0.099 \\
    &  55 &  50 &   35.78 & $[0.000,\,0.208]$      & 0.208 \\
    &  55 & 100 &   37.42 & $[0.185,\,0.435]$      & 0.435 \\
    &  55 & 200 &   39.30 & $[0.666,\,0.916]$      & 0.915 \\
    &  62 &  25 &   25.56 & $[0.516,\,0.767]$      & 0.563 \\
    &  62 &  50 &   27.57 & $[1.179,\,1.430]$      & 1.150 \\
    &  62 &  60 &   28.36 & $[1.415,\,1.666]$      & 1.388 \\
    &  62 &  70 &   28.85 & $[1.644,\,1.893]$      & 1.628 \\
    &  62 & 100 &   30.31 & $[2.308,\,2.558]$      & 2.361 \\
    &  62 & 200 &   33.01 & $[4.654,\,4.904]$      & 4.912 \\
    &  65 &  25 &   21.95 & $[3.623,\,3.891]$      & \textit{invalid} ($\rho_4^{Q}{=}3.32$) \\
    &  65 &  50 &   24.34 & $[5.249,\,5.500]$      & \textit{invalid} ($\rho_4^{Q}{=}1.48$) \\
    &  65 &  60 &   25.01 & $[5.886,\,6.129]$      & \textit{invalid} ($\rho_4^{Q}{=}1.19$) \\
    &  65 &  70 &   25.67 & $[6.531,\,6.767]$      & 4.999 \\
    &  65 &  80 &   26.30 & $[7.167,\,7.415]$      & 5.712 \\
    &  65 &  90 &   26.81 & $[7.818,\,8.072]$      & 6.428 \\
    &  65 & 100 &   27.25 & $[8.482,\,8.737]$      & 7.147 \\
    &  65 & 120 &   27.99 & $[9.839,\,10.090]$     & 8.596 \\
    &  65 & 140 &   28.72 & $[11.228,\,11.472]$    & 10.060 \\
    &  65 & 200 &   30.05 & $[15.522,\,15.769]$    & 14.552 \\
    &  67 &  25 &   17.94 & $[13.360,\,13.657]$    & \textit{invalid} ($\rho_4^{Q}{=}3.46$) \\
    &  67 &  50 &   20.39 & $[20.517,\,20.746]$    & \textit{invalid} ($\rho_4^{Q}{=}1.28$) \\
    &  67 &  80 &   22.50 & $[28.670,\,28.950]$    & 24.433 \\
    &  67 &  90 &   23.16 & $[31.387,\,31.670]$    & 27.260 \\
    &  67 & 100 &   23.67 & $[34.120,\,34.392]$    & 30.085 \\
    &  67 & 120 &   24.63 & $[39.619,\,39.854]$    & 35.739 \\
    &  67 & 140 &   25.35 & $[45.108,\,45.350]$    & 41.411 \\
    &  67 & 200 &   26.93 & $[61.908,\,62.120]$    & 58.631 \\
    &  70 &  25 &   14.70 & $[52.702,\,53.019]$    & \textit{invalid} ($\rho_4^{P}{=}2.26$) \\
    &  70 &  50 &   17.44 & $[83.289,\,83.470]$    & \textit{invalid} ($\rho_4^{P}{=}1.15$) \\
    &  70 & 100 &   20.97 & $[140.112,\,140.393]$  & 132.911 \\
    &  70 & 200 &   24.54 & $[251.533,\,251.653]$  & 245.964 \\
  \midrule
  5 &  55 &  25 &   28.81 & $[0.000,\,0.099]$      & 0.099 \\
    &  55 &  50 &   29.63 & $[0.000,\,0.208]$      & 0.208 \\
    &  55 & 100 &   31.20 & $[0.185,\,0.435]$      & 0.435 \\
    &  55 & 200 &   32.72 & $[0.666,\,0.916]$      & 0.915 \\
    &  62 &  25 &   26.17 & $[0.516,\,0.767]$      & 0.563 \\
    &  62 &  50 &   28.28 & $[1.179,\,1.430]$      & 1.150 \\
    &  62 & 100 &   31.21 & $[2.308,\,2.558]$      & 2.361 \\
    &  62 & 200 &   33.94 & $[4.654,\,4.904]$      & 4.912 \\
    &  65 &  25 &   24.37 & $[3.623,\,3.891]$      & \textit{invalid} ($\rho_4^{Q}{=}3.32$) \\
    &  65 &  50 &   26.86 & $[5.249,\,5.500]$      & \textit{invalid} ($\rho_4^{Q}{=}1.48$) \\
    &  65 & 100 &   29.66 & $[8.482,\,8.737]$      & 7.147 \\
    &  65 & 200 &   32.50 & $[15.522,\,15.769]$    & 14.552 \\
    &  67 &  25 &   22.17 & $[13.360,\,13.657]$    & \textit{invalid} ($\rho_4^{Q}{=}3.46$) \\
    &  67 &  50 &   24.53 & $[20.517,\,20.746]$    & \textit{invalid} ($\rho_4^{Q}{=}1.28$) \\
    &  67 & 100 &   27.70 & $[34.120,\,34.392]$    & 30.085 \\
    &  67 & 200 &   30.67 & $[61.908,\,62.120]$    & 58.631 \\
    &  70 &  25 &   20.05 & $[52.702,\,53.019]$    & \textit{invalid} ($\rho_4^{P}{=}2.26$) \\
    &  70 &  50 &   22.64 & $[83.289,\,83.470]$    & \textit{invalid} ($\rho_4^{P}{=}1.15$) \\
    &  70 & 100 &   25.88 & $[140.112,\,140.393]$  & 132.911 \\
    &  70 & 200 &   29.24 & $[251.533,\,251.653]$  & 245.964 \\
  \midrule
  7 &  55 &  25 &   29.25 & $[0.000,\,0.099]$      & 0.099 \\
    &  55 &  50 &   29.67 & $[0.000,\,0.208]$      & 0.208 \\
    &  55 & 100 &   30.73 & $[0.185,\,0.435]$      & 0.435 \\
    &  55 & 200 &   31.74 & $[0.666,\,0.916]$      & 0.915 \\
    &  62 &  25 &   26.42 & $[0.516,\,0.767]$      & 0.563 \\
    &  62 &  50 &   28.15 & $[1.179,\,1.430]$      & 1.150 \\
    &  62 & 100 &   30.47 & $[2.308,\,2.558]$      & 2.361 \\
    &  62 & 200 &   32.79 & $[4.654,\,4.904]$      & 4.912 \\
    &  65 &  25 &   25.73 & $[3.623,\,3.891]$      & \textit{invalid} ($\rho_4^{Q}{=}3.32$) \\
    &  65 &  50 &   27.96 & $[5.249,\,5.500]$      & \textit{invalid} ($\rho_4^{Q}{=}1.48$) \\
    &  65 & 100 &   30.43 & $[8.482,\,8.737]$      & 7.147 \\
    &  65 & 200 &   33.00 & $[15.522,\,15.769]$    & 14.552 \\
    &  67 &  25 &   25.02 & $[13.360,\,13.657]$    & \textit{invalid} ($\rho_4^{Q}{=}3.46$) \\
    &  67 &  50 &   27.37 & $[20.517,\,20.746]$    & \textit{invalid} ($\rho_4^{Q}{=}1.28$) \\
    &  67 & 100 &   30.17 & $[34.120,\,34.392]$    & 30.085 \\
    &  67 & 200 &   32.95 & $[61.908,\,62.120]$    & 58.631 \\
    &  70 &  25 &   23.51 & $[52.702,\,53.019]$    & \textit{invalid} ($\rho_4^{P}{=}2.26$) \\
    &  70 &  50 &   25.97 & $[83.289,\,83.470]$    & \textit{invalid} ($\rho_4^{P}{=}1.15$) \\
    &  70 & 100 &   29.19 & $[140.112,\,140.393]$  & 132.911 \\
    &  70 & 200 &   32.23 & $[251.533,\,251.653]$  & 245.964 \\
  \midrule
  9 &  55 &  25 &   31.27 & $[0.000,\,0.099]$      & 0.099 \\
    &  55 &  50 &   31.25 & $[0.000,\,0.208]$      & 0.208 \\
    &  55 & 100 &   31.69 & $[0.185,\,0.435]$      & 0.435 \\
    &  55 & 200 &   32.19 & $[0.666,\,0.916]$      & 0.915 \\
    &  62 &  25 &   27.50 & $[0.516,\,0.767]$      & 0.563 \\
    &  62 &  50 &   28.47 & $[1.179,\,1.430]$      & 1.150 \\
    &  62 & 100 &   29.87 & $[2.308,\,2.558]$      & 2.361 \\
    &  62 & 200 &   31.47 & $[4.654,\,4.904]$      & 4.912 \\
    &  65 &  25 &   26.39 & $[3.623,\,3.891]$      & \textit{invalid} ($\rho_4^{Q}{=}3.32$) \\
    &  65 &  50 &   27.89 & $[5.249,\,5.500]$      & \textit{invalid} ($\rho_4^{Q}{=}1.48$) \\
    &  65 & 100 &   29.73 & $[8.482,\,8.737]$      & 7.147 \\
    &  65 & 200 &   31.58 & $[15.522,\,15.769]$    & 14.552 \\
    &  67 &  25 &   26.09 & $[13.360,\,13.657]$    & \textit{invalid} ($\rho_4^{Q}{=}3.46$) \\
    &  67 &  50 &   27.78 & $[20.517,\,20.746]$    & \textit{invalid} ($\rho_4^{Q}{=}1.28$) \\
    &  67 & 100 &   29.93 & $[34.120,\,34.392]$    & 30.085 \\
    &  67 & 200 &   32.34 & $[61.908,\,62.120]$    & 58.631 \\
    &  70 &  25 &   25.30 & $[52.702,\,53.019]$    & \textit{invalid} ($\rho_4^{P}{=}2.26$) \\
    &  70 &  50 &   27.54 & $[83.289,\,83.470]$    & \textit{invalid} ($\rho_4^{P}{=}1.15$) \\
    &  70 & 100 &   30.32 & $[140.112,\,140.393]$  & 132.911 \\
    &  70 & 200 &   32.90 & $[251.533,\,251.653]$  & 245.964 \\
\end{longtable}
\end{footnotesize}

\begin{table}[h]
\centering
\caption{CelebA-64, $B = 100$, three public re-noising steps, $\delta = 10^{-6}$, $N_{\mathrm{out}} = 50{,}000$. FID is mean $\pm$ standard deviation over three generation seeds.}
\label{tab:celeba_seeds}
\small
\begin{tabular}{crrrr}
\toprule
Handoff & Edgeworth estimate & PLD lower & Certified PLD upper & FID \\
\midrule
55\% & 0.43 & 0.19 & 0.44 & $37.39 \pm 0.05$ \\
62\% & 2.36 & 2.31 & 2.56 & $30.45 \pm 0.17$ \\
65\% & 7.14 & 8.48 & 8.74 & $27.31 \pm 0.07$ \\
67\% & 30.09 & 34.12 & 34.39 & $23.75 \pm 0.08$ \\
70\% & 132.91 & 140.11 & 140.39 & $20.98 \pm 0.05$ \\
\bottomrule
\end{tabular}
\end{table}

\begin{table}[t]
  \centering
  \small
  \setlength{\tabcolsep}{8pt}
  \begin{tabular}{crrr}
    \toprule
    \textbf{Handoff (\%)}
      & Edgeworth estimate
      & CLT estimate
      & Certified PLD upper \\
    \midrule
    55 &      0.299 &        0.296 &   0.299 \\
    62 &      1.683 &        1.388 &   1.691 \\
    65 &      5.331 &        4.313 &   6.289 \\
    67 &     22.982 &       45.101 &  25.696 \\
    70 &    104.035 & 25{,}905.494 & 108.444 \\
    \bottomrule
  \end{tabular}
  \caption{%
    Edgeworth and CLT estimates of $\varepsilon$ against the certified PLD upper bound
    (CIFAR-10, $B = 25$, three re-noising steps, $\delta = 10^{-5}$). Neither approximation
    is a bound. The CLT estimate understates the certified value by 18\% and 31\% at the
    62\% and 65\% blocks and overstates it by more than 200 times at the 70\% block. The
    Edgeworth estimate stays within 1\% for $\varepsilon \lesssim 2$ and understates it by
    15\%, 11\%, and 4\% at the 65\%, 67\%, and 70\% blocks.
  }
  \label{tab:edgeworth_vs_clt}
\end{table}

\section{Selection of the Privacy Parameters $\varepsilon$ and $\delta$}
\label{app:eps-delta}

For a fixed sampler, Propositions~\ref{prop:output_subsampling} and~\ref{prop:step_subsampling} determine a complete privacy profile
$\delta_{\mathcal M}(\varepsilon)$: a practitioner picks a target
$\delta_{\mathrm{target}}$ and reports the smallest $\varepsilon$ with
$\delta_{\mathcal M}(\varepsilon)\leq\delta_{\mathrm{target}}$, so $\delta$ is
not tuned only through $\varepsilon$ --- the pair describes one fixed mechanism.
Arbitrarily small $\varepsilon$ and $\delta$ cannot coexist at fixed utility;
improving both requires changing the mechanism (lower sampling rate, fewer
private steps or released outputs, smaller clipping bound, more noise, or an
earlier handoff).

We used $\delta=10^{-5}\approx1/n$ for CIFAR-10 to match common DP
generative-model comparisons; $1/n^2=4\times10^{-10}$ is indeed stricter, and we
report both in Table~\ref{tab:delta-sweep}.

\begin{table}[h]
\centering
\caption{Certified RDP $\varepsilon$ for CIFAR-10 output-wise subsampled blocks
at $\delta=10^{-5}$ and at the stricter $\delta=4\times10^{-10}=1/n^2$
($B=25$, $N_{\mathrm{out}}=50{,}000$).}
\label{tab:delta-sweep}
\begin{tabular}{lrr}
\toprule
Configuration & RDP $\varepsilon$, $\delta=10^{-5}$ & RDP $\varepsilon$, $\delta=4\times10^{-10}$ \\
\midrule
55\%, $B=25$ & 0.36 & 0.71 \\
62\%, $B=25$ & 2.63 & 5.16 \\
65\%, $B=25$ & 7.81 & 12.87 \\
\bottomrule
\end{tabular}
\end{table}

\section{Computing theoretical ASR}\label{sec:ASR}

Following \citet{triastcyn2020bayesian}, we calculate the theoretical ASR as follows. Suppose an adversary observed an $\epsilon$-DP release $o$ and tried to distinguish between two candidate datasets $\mathcal{D}, \mathcal{D'}$ with a uniform prior $p(\mathcal{D})=p(\mathcal{D'})=0.5$. Without loss of generality, let $\mathcal{D}$ be the correct dataset, then the adversary will infer the probability $p(\mathcal{D}|o)$
as
$p(\mathcal{D}|o)=\frac{p(o|\mathcal{D})p(\mathcal{D})}{p(o|\mathcal{D})p(\mathcal{D})+p(o|\mathcal{D'})p(\mathcal{D'})} \leq \frac{p(o|\mathcal{D})}{p(o|\mathcal{D})+ \exp(-\epsilon)p(o|\mathcal{D})} = \frac{1}{1+\exp(-\epsilon)}$. For $(\varepsilon,\delta)$-DP the bound becomes $\frac{e^{\varepsilon}+\delta}{1+e^{\varepsilon}}$, which at $\delta = 10^{-5}$ changes no entry of \tabref{MIA} at the reported precision.
So, for $\epsilon=1$ the upperbound on the membership inference success probability is $73.1\%$, for $\epsilon=5$, it is $99.33\%$, and for $\epsilon=10$, it is  $99.995\%$. Furthermore, the success probability of MIAs typically similar or lower than the theoretical success rate as shown in \citet{lowy2024largeEps}.

\begin{table}[ht]
\centering
\caption{Membership inference. Theoretical: ASR implied by our $\varepsilon$ for the subsampled empirical sampler at rate $q$. Numerical: ASR of the attack of \citet{duan2023are} on a neural denoiser trained on all of CIFAR-10. Both stop at time $t$ of a $T = 1000$ step discretization. The neural denoiser is not subsampled, so the theoretical rows are not a bound on it.}
\scalebox{0.8}{\begin{tabular}{lcccc}
\hline
$t$ & $T$ & $\tfrac{3}{4}T$ & $\tfrac{1}{2}T$ & $\tfrac{1}{4}T$ \\
\hline
Cosine similarity (Fig.~\ref{fig:denoiser_comparison})
& 1.000 & 0.999 & 0.956 & 0.845 \\
Theoretical ASR at $q = 0.5$
& 0.5019 & 0.5204 & 0.6909 & 0.9999 \\
Theoretical ASR at $q = 0.01$
& 0.5000 & 0.5002 & 0.5031 & 0.6927 \\
Theoretical ASR at $q = 0.005$
& 0.5000 & 0.5001 & 0.5012 & 0.6371 \\
Numerical ASR
& 0.5005 $\pm$ 0.0025 & 0.5001 $\pm$ 0.0011 & 0.5072 $\pm$ 0.0007 & 0.5096 $\pm$ 0.0026 \\
\hline
\end{tabular}\label{tab:MIA}}
\end{table}

\section{Broader Impact Statement}
\label{app:impact_statement}

This work studies privacy leakage that can arise during diffusion-based sampling and provides tools for quantifying how individual training records may influence generated outputs. Its primary positive impact is to support more privacy-aware deployment of generative models, especially in settings where training data may contain sensitive or proprietary information. By shifting attention from only differentially private training to sampling-time privacy accounting, the proposed framework may help practitioners design generative pipelines that better balance privacy and utility, for example through subsampling, early stopping, or hybrid use of public denoisers. This can make formal privacy analysis more accessible for large-scale diffusion systems where fully private training remains costly or impractical.

At the same time, the work has limitations and possible risks. The formal guarantees are derived for an empirical denoiser with tractable sensitivity, while modern deployed systems typically use neural denoisers whose exact data sensitivity remains difficult to characterize. As a result, the analysis should not be interpreted as a complete privacy certificate for arbitrary diffusion models. If misused or overstated, sampling-time privacy estimates could give a false sense of security. Moreover, improving the privacy--utility tradeoff of generative models may also make it easier to deploy high-quality synthetic data systems in sensitive domains, where failures could still expose private information or amplify biases present in the underlying data.

We therefore view this work as a step toward, rather than a substitute for, end-to-end privacy evaluation. Responsible use should combine sampling-time accounting with empirical privacy audits, membership-inference evaluation, dataset governance, and clear reporting of assumptions. More broadly, our analysis highlights that privacy in generative modeling is shaped not only by training, but also by how samples are produced, released, and reused.

\end{document}